\documentclass[journal]{IEEEtran}
\usepackage{cite}
\usepackage{comment}
\usepackage{floatrow}
\usepackage{booktabs}
\newfloatcommand{capbtabbox}{table}[][\FBwidth]

\usepackage{subfiles} 
\usepackage{dsfont}
\ifCLASSINFOpdf
   \usepackage[pdftex]{graphicx}
\else
\fi
\usepackage{amsmath,amssymb,amsfonts,amsthm}
\usepackage{array}

\usepackage[caption=false,font=footnotesize]{subfig}
\usepackage{url}
\usepackage{caption}
\usepackage{comment,enumerate}
\usepackage{bm}
\usepackage{booktabs, multirow}
\usepackage[ruled,vlined]{algorithm2e}
\usepackage{tikz}
\usetikzlibrary{bayesnet}
\usetikzlibrary{arrows}

\newtheorem{theorem}{Theorem}
\newtheorem{proposition}[theorem]{Proposition}
\newtheorem{lemma}[theorem]{Lemma}

\newtheorem{problem}{Problem}

\DeclareMathOperator*{\vol}{vol}
\DeclareMathOperator*{\argmin}{argmin}

\DeclareMathOperator*{\rank}{rank}
\DeclareMathOperator*{\tr}{tr}
\DeclareMathOperator*{\Cov}{Cov}

\DeclareMathOperator*{\vecsym}{vsym}

\newcommand{\T}{{\mathbb{T}}}
\newcommand{\R}{{\mathbb{R}}}
\newcommand{\E}{{\mathbb{E}}}
\newcommand{\bZ}{{\bm{Z}}}
\newcommand{\bX}{{\bm{X}}}
\newcommand{\deltat}{{\Delta t}}

\def\d{\,\mathrm{d}}
\newcommand{\cX}{{\mathcal{X}}}
\newcommand{\cZ}{{\mathcal{Z}}}

\let\sectionlabel\label

\newcommand{\mathscalebox}[2]{%
	\text{\scalebox{#1}{$\displaystyle #2$}}}

\begin{document}

%
\title{Identifying Latent Stochastic Differential Equations}
%
%
%

\author{Ali~Hasan*,
        Jo\~ao M. Pereira*,
        Sina~Farsiu, 
        and Vahid~Tarokh 
\thanks{Manuscript submitted May 5, 2021. This work was supported in part by Office of Naval Research Grant No. N00014-18-1-224.}
\thanks{A. Hasan and S. Farsiu are with the Department
of Biomedical Engineering at Duke University, Durham,
NC \\ 
Correspondence e-mail: ali.hasan@duke.edu}
\thanks{J. M. Pereira is with the Oden Institute for Computational Engineering and Sciences at University of Texas at Austin}
\thanks{V. Tarokh is with the Department of Electrical and Computer Engineering at Duke University}

\thanks{}
}
\maketitle

\begin{abstract}
We present a method for learning latent stochastic differential equations (SDEs) from high dimensional time series data.
Given a high-dimensional time series generated from a lower dimensional latent unknown It\^{o} process, the proposed method learns the mapping from ambient to latent space, and the underlying SDE coefficients, through a self-supervised learning approach.
Using the framework of variational autoencoders, we consider a conditional generative model for the data based on the Euler-Maruyama approximation of SDE solutions.
Furthermore, we use recent results on identifiability of latent variable models to show that the proposed model can recover not only the underlying SDE coefficients, but also the original latent variables, up to an isometry, in the limit of infinite data.
We validate the method through several simulated video processing tasks, where the underlying SDE is known, and through real world datasets.
\end{abstract}


%
\IEEEpeerreviewmaketitle

\section{Introduction}
%
%
%
%
\IEEEPARstart{V}{ariational} auto-encoders (VAEs) are a widely used tool to learn lower-dimensional latent representations of high-dimensional data.
However, the learned latent representations often lack interpretability, and it is challenging to extract relevant information from the representation of the dataset in the latent space. 
In particular, when the high-dimensional data is governed by unknown and lower-dimensional dynamics, arising, for instance, from unknown physical or biological interactions, the latent space representation often fails to bring insight on these dynamics. 

To address this shortcoming, we propose a VAE-based framework for recovering latent dynamics governed by stochastic differential equations (SDEs). SDEs are a generalization of ordinary differential equations, that contain both a deterministic term, denoted by drift coefficient, and a stochastic term, denoted by diffusion coefficient.
SDEs are often used to study stochastic processes, with applications ranging from modeling physical and biological phenomena to financial markets.
Moreover, their properties have been extensively studied in the fields of probability and statistics, and a rich set of tools for analyzing these have been developed.
However, most tools are limited to lower dimensional settings, which further motivates recovering lower dimensional latent representations of the data. 

To define the problem, suppose we observe a high-dimensional time-series $\{X_t\}_{t\in \bm{T}}$, for which there exists a 
unknown latent representation $\{Z_t\}_{t\in \bm{T}}$ which is governed by an SDE, with drift and diffusion coefficients that are also unknown. 
More specifically, the latent representation is defined by an injective function $f$, which we denote by latent mapping, such that 
\begin{equation}\label{eq:fplusnoise}
X_t = f(Z_t) + \epsilon_t,\quad t\in\bm{T}
\end{equation}
where the noise terms $\{\epsilon_t\}_{t\in \bm{T}}$ are i.i.d. and independent of $\{Z_t\}_{t\in \bm{T}}$. In this paper, we propose a VAE-based model for recovering both the latent mapping and the coefficients of the SDE that governs $Z_t$. 

\begin{figure*}
	\centering
	\subfloat[Yellow ball moving according to a 2D Ornstein-Uhlenbeck process;]{\includegraphics[width=\textwidth]{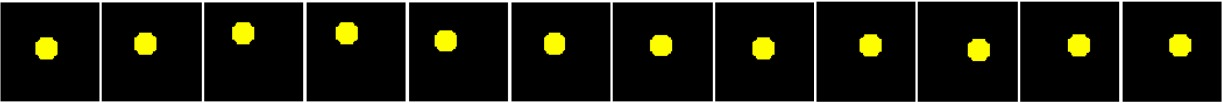}%
	\label{fig:12_balls}}\\
	\subfloat[Comparison between the true centers of the ball and the latent representation learned by the VAE at different frames of the video;]{
		\includegraphics[width=0.48\textwidth]{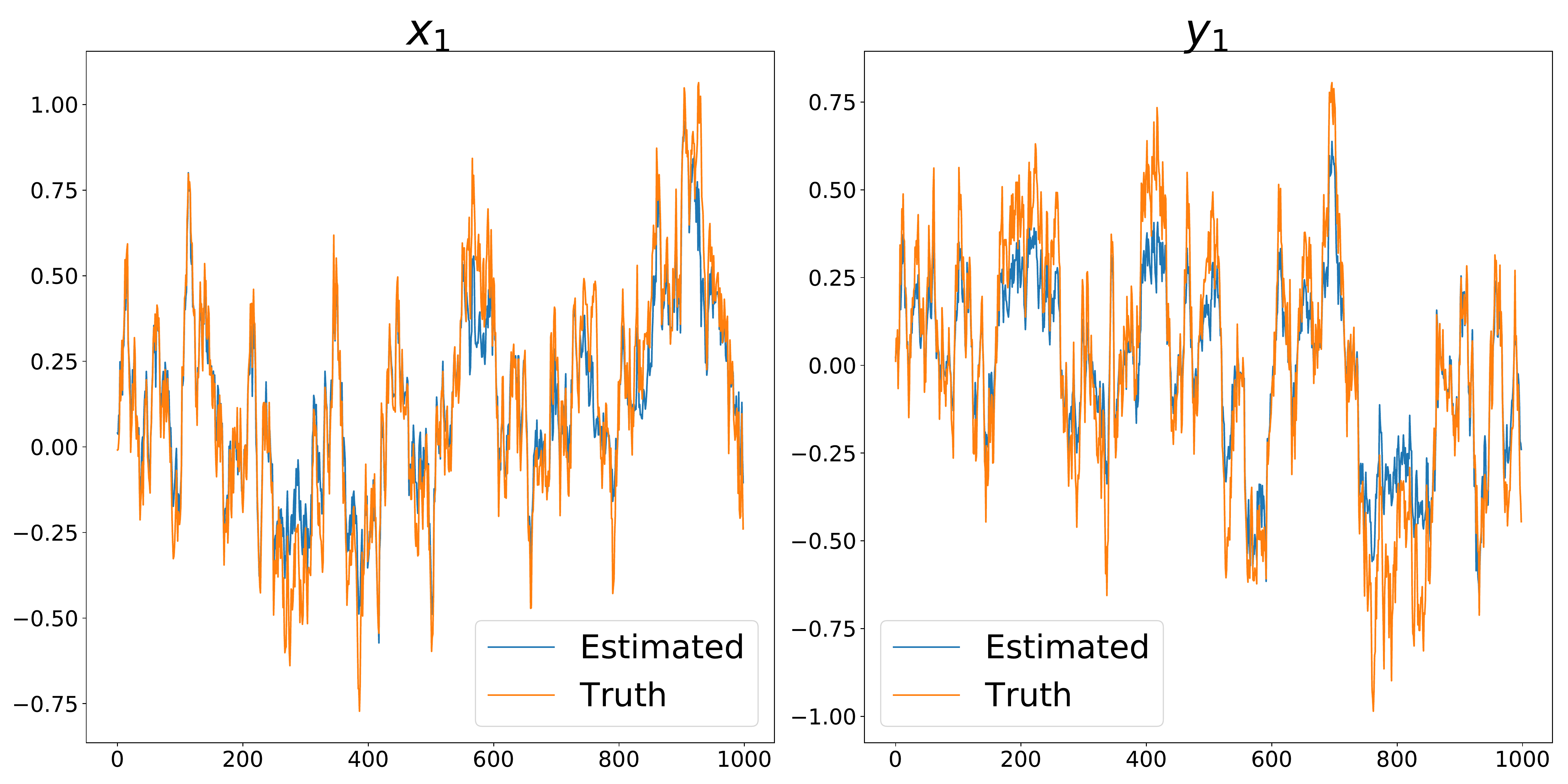}
		\label{fig:example_coord_learned}
    }   \hfill
	\subfloat[Comparison between the true drift coefficient of the underlying SDE, the drift coefficient learnt by SDE--VAE, and the drift coefficient learnt by an oracle that knows the true latent variables. The difficulty in estimation is seen even in the oracle estimate;]{
		\includegraphics[width=0.48\textwidth]{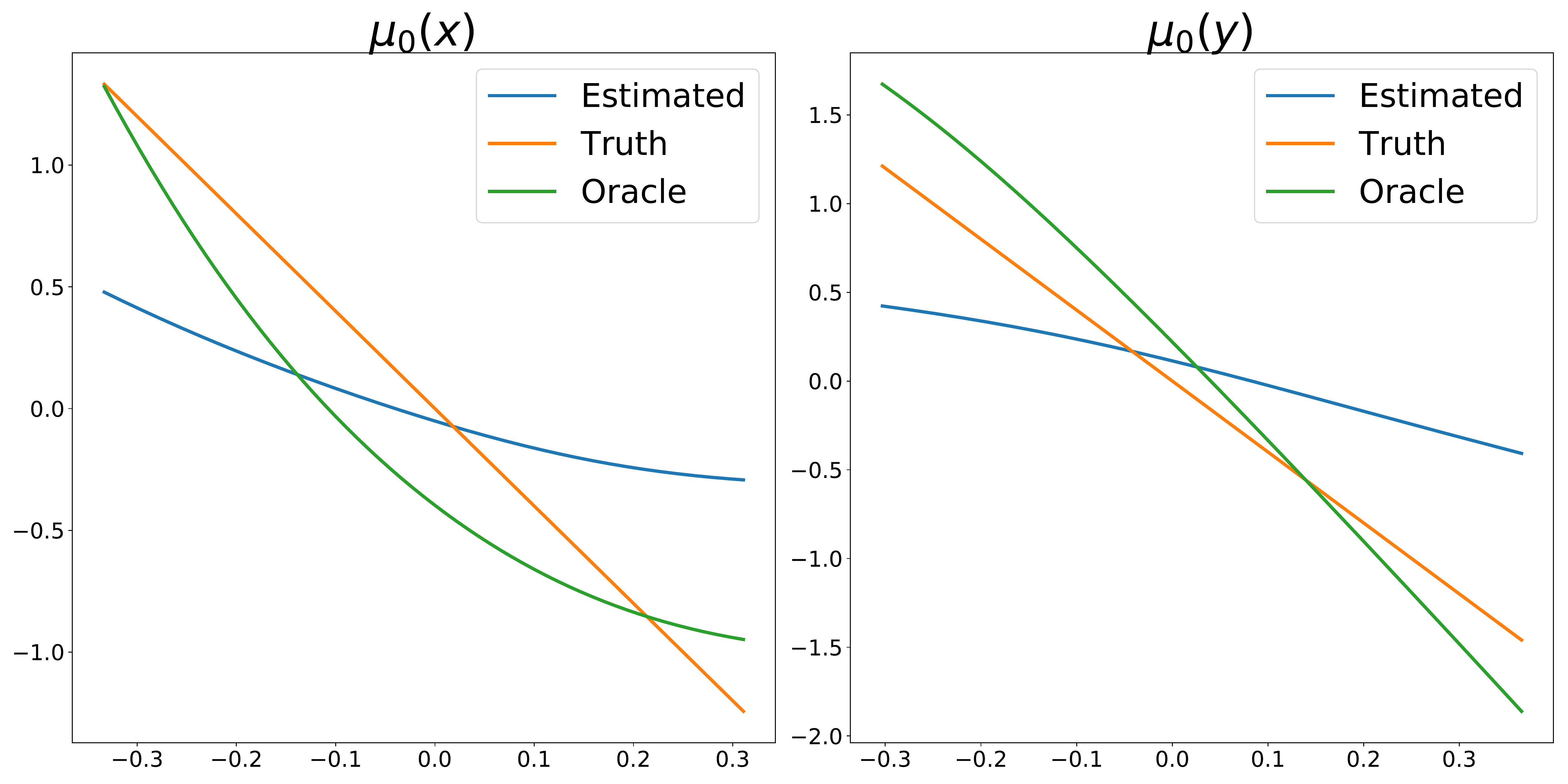}
	\label{fig:example_mu_learned}
	}   \hfill
	\caption{From a video of a yellow ball moving in the plane, according to a 2D Ornstein-Uhlenbeck process (Fig.~\ref{fig:12_balls}), the proposed model learns that the relevant latent representation of each frame are the $x$ and $y$ coordinates of the ball (Fig.~\ref{fig:example_coord_learned}), and learns the drift coefficient of the SDE (Fig.~\ref{fig:example_mu_learned}).}\label{fig:SDE_VAE_example}
\end{figure*}
We are also concerned with identifiability. Since the latent representation is unknown, applying any one-to-one mapping to $Z_t$ yields another latent representation $\tilde Z_t$ of $X_t$, with different latent mapping and latent SDE dynamics. To pick out one latent representation, up to equivalence by one-to-one transformations on the latent space, we propose the following two-fold approach.
\begin{enumerate}[(i)]
    \item \label{item:1} Under some conditions over the coefficients of the SDE that governs $Z_t$, we show that there exists another latent representation $\tilde Z_t$ of $X_t$, such that $\tilde Z_t$ is governed by an SDE with an \emph{isotropic} diffusion coefficient (Theorem \ref{thm:canonical_sigma_}). 
    \item \label{item:2} We prove that the method proposed in this paper, in the limit of infinite data, is able to recover, up to an isometry, the mapping from $X_t$ to $\tilde Z_t$ and the drift coefficient of the SDE that governs $\tilde Z_t$ (Theorems \ref{thm:identifiability_main}, \ref{thm:VAE_convergence}).
\end{enumerate}

%
By assuming the diffusion coefficient is isotropic, our approach has an easier task of learning the latent dynamics, since the diffusion coefficient does not need to be estimated. An example of the proposed method is presented in Fig.~\ref{fig:SDE_VAE_example}.

Our paper is organized as follows. First, we present an overview of previous work. Then we review the notion of SDEs, develop a generative model to study latent SDEs, and present the VAE framework that enables learning of the proposed model. Followingly, we show that the VAE proposed recovers the true model parameters up to an isometry and give some practical considerations on the method presented. 
Finally, we test the proposed method in several synthetic and real world video datasets, governed by lower dimensional SDE dynamics, and present a brief discussion on the results.

\section{Related Work}\label{sec:related_work}

Previous work in learning SDEs has been mostly focused on lower-dimensional data. 
Classical approaches assume fixed drift and diffusion coefficients with parameters that need to be estimated \cite{iacus2009simulation}. 
In \cite{rudy2017data}, a method is proposed where the terms of the Fokker-Planck equation are estimated using sparse regression with a predefined dictionary of functions, and in \cite{boninsegna2018sparse} a similar idea is applied to the Kramers-Moyal expansion of the SDE.
In \cite{jia2019neural}, the authors describe a method that makes use of a SDE driven by a counting process, while
\cite{yildiz2018learning} describes a method for recovering an SDE using Gaussian processes.
The statistical model we introduce for learning latent SDEs is similar to a Hidden Markov Model (HMM) with a complicated emission model. For this problem, spectral methods \cite{hsu2012spectral}, and extensions with non-parametric emission models \cite{kandasamy2016learning, song2010hilbert, song2014nonparametric}, have been proposed.

The work that resembles the most our contribution is \cite{duncker2019learning}, where a method is presented for uncovering the latent SDE for high-dimensional data using Gaussian processes.
However, the method assumes that there is an intermediate feature space, such that the map from ambient space to feature space is known, and the map from feature space to latent space is linear and unknown. Its applicability is therefore limited when it is not clear what features of the data should be considered.

Regarding work that involves neural networks, \cite{tzen2019neural} describes a variational inference scheme for SDEs using neural networks, and in \cite{yildiz2019ode2vae}, a method using variational auto-encoders is presented to recover latent second-order ordinary differential equations from data, but the dynamics are assumed to be governed by a deterministic ODE. 
Other related works that exploit knowledge the existence of SDEs are \cite{li2020scalable}, where the adjoint sensitivity method generalized for backpropogating through an SDE solver to train neural networks and \cite{ha2018adaptive} which describes a method for using an autoencoder with path integrals in control scenarios.
However, none are interested in recovering an underlying SDE or analyzing what SDE was recovered. 

Finally, in the case of image/video data, recent works in stochastic video prediction \cite{babaeizadeh2018stochastic,Kumar2020VideoFlow} describe methods for stochastic predictions of video. While these are favorable on reproducing the dynamics of the observed data, the latent variables lack interpretability. 
In \cite{hsieh2018learning}, a method based on recurrent neural networks is presented that decomposes the latent space and promotes disentanglement, in an effort to provide more meaningful features in the latent space.

In all of the related work, none address the problem of recovering an underlying SDE given high dimensional measurements, balancing both interpretability and efficacy in modeling complex data sets.
The proposed method aims to fill this gap.

\section{Stochastic Differential Equations}
Here we review the definition of SDE. For a time interval $\T = [0,T]$, let $\{W_t\}_{t\in \T}$ be a $d$-dimensional Wiener process.
We say the stochastic process $\{Z_t\}_{t\in \T}\in \R^d$ is a solution to the It\^o SDE
\begin{equation}\label{eq:SDE_def}
\d Z_t = \mu(Z_t, t) \d t + \sigma(Z_t, t) \d W_t,
\end{equation}
if $Z_0$ is independent of the $\sigma$-algebra generated by $W_t$, and
\begin{equation}\label{eq:SDE_def_int}
Z_{T} = Z_0 + \int_0^T \mu(Z_t, t) \d t + \int_0^T  \sigma(Z_t, t) \d W_t.
\end{equation}
Here we denote the \emph{drift coefficient} by $\mu:\R^d\times \T\to \R^d$, the \emph{diffusion coefficient} by $\sigma:\R^d\times \T\to \R^{d\times d}$ and the second integral in \eqref{eq:SDE_def_int} is the It\^o stochastic integral \cite{oksendal2003stochastic}. When the coefficients are globally Lipschitz, that is,
\begin{equation}\label{eq:mu_sigma_Lipschitz}
\begin{split}
\|\mu(x,t)-\mu(y,t)\| + \|\sigma(x,t)-\sigma(y,t)\|\le D\|x-y\|\quad  \\ \forall x,y\in\R^d, t\in \T,
\end{split}
\end{equation}
for some constant $D>0$, there exists a unique $t$-continuous strong solution to \eqref{eq:SDE_def} \cite[Theorem 5.2.1]{oksendal2003stochastic}. Finally, throughout the paper we can assume $\sigma(z, t)$ is a symmetric positive semi-definite matrix for all $z\in \R^d$ and $t\in \T$, which follows from \cite[Theorem 7.3.3]{oksendal2003stochastic}.

For ease of exposition, we present our main results for SDEs with time independent coefficients, and extend the results to time-dependent coefficients in Section \ref{sec:practical_considerations}.B. 




\section{Problem Definition}

In this paper, we consider a high-dimensional stochastic process $\{X_t\}_{t \in \T} \in \R^n$, which has a latent representation $\{Z_t\}_{t \in \T}\in \R^d$, with $n\ge d$, as defined in \eqref{eq:fplusnoise}.
Moreover, $Z_t$ is governed by an SDE, with drift coefficient $\mu:\R^d\to \R^d$ and diffusion coefficient $\sigma:\R^d\to \R^{d\times d}$. The aim of this paper is to recover the latent mapping $f$ and the coefficients of the SDE that governs $Z_t$ ($\mu$ and $\sigma$) from $X_t$. 
We consider the following problem.
\begin{problem}\label{prob:identifiability}
Find $(f, \mu, \sigma)$ such that \eqref{eq:fplusnoise} holds, and $\{Z_t\}_{t \in \T}$ is a solution to the SDE with drift and diffusion coefficients $\mu$ and $\sigma$, respectively.
\end{problem}

By definition, the latent space is unknown, so any one-to-one transformation of the latent space cannot be recovered from the observed data. Therefore there is an inherent ambiguity of one-to-one functions for Problem \ref{prob:identifiability}, which we formalize as follows.

\begin{proposition}\label{prop:1}
Consider the equivalence relation,
    \begin{equation}
    (f, \mu, \sigma) \sim (\tilde f,\tilde \mu, \tilde \sigma),
    \end{equation}
    if there is an invertible function $g:\R^d \to \R^d$ such that
    \begin{itemize}
    \item For any solution $Y_t$ of \eqref{eq:SDE_def}, $g(Y_t)$ is a solution to \eqref{eq:SDE_def} with drift and diffusion coefficients $\tilde \mu$ and $\tilde \sigma$, respectively%
    \footnote{It\^o's Lemma \cite{kunita1967square} implies that if $Y_t$ is the solution of an SDE, then $g(Y_t)$ is also the solution of another SDE, for which the drift and diffusion coefficients, $\tilde \mu$ and $\tilde \sigma$, can be explicitly written in terms of $\mu$, $\sigma$ and $g$.};
    \item $\tilde f(z) = (f \circ g^{-1})(z)\quad  \forall z\in \R^d$.
    \end{itemize}
	
	Then if $(f, \mu, \sigma)$ is a solution to Problem \ref{prob:identifiability} and $(\tilde f,\tilde \mu, \tilde \sigma)\sim (f,\mu, \sigma)$, then $(\tilde f, \tilde \mu, \tilde \sigma)$ is also a solution to Problem \ref{prob:identifiability}. In particular, we can only recover $(f,\mu, \sigma)$ up to its equivalence class.
\end{proposition}
\begin{proof}
If $(f, \mu, \sigma) \sim (\tilde f,\tilde \mu, \tilde \sigma)$, then there exists an invertible function $g$ such that $\tilde \mu$ and $\tilde \sigma$ are the drift and diffusion coefficients of the SDE that governs $\tilde Z_t = g(Z_t)$. We have $\tilde f = f \circ g^{-1}$, which implies,
\begin{equation*}
\tilde f(\tilde Z_t) = f(g^{-1}(g(Z_t))) = f(Z_t),
\end{equation*} and \eqref{eq:fplusnoise} also holds with $f$ and $Z_t$ replaced by $\tilde f$ and $\tilde Z_t$, respectively, thus $(\tilde f,\tilde \mu, \tilde \sigma)$ is also a solution to Problem \ref{prob:identifiability}.
\end{proof}
Since we can only recover $(f, \mu, \sigma)$ up to its equivalence class, we should focus on recovering an element of the equivalence class which is easier to describe. The following theorem achieves that: under some conditions on $\mu$ and $\sigma$, there is other element $(\tilde f, \tilde \mu, \tilde \sigma)$ in the same equivalence class of $(f, \mu, \sigma)$ for which $\tilde \sigma$ is \emph{isotropic}, that is, $\tilde \sigma(y) = I_d$ for all $y\in \R^d$, where $I_d$ is the identity matrix of size $d$.

\begin{theorem}\label{thm:canonical_sigma_}
	Suppose that $(f, \mu, \sigma)$ is a solution to Problem \ref{prob:identifiability}, and that the following conditions are satisfied:
	\begin{enumerate}[(\ref{thm:canonical_sigma_}.i)]
		\item $\mu$ and $\sigma$ are globally Lipschitz as in \eqref{eq:mu_sigma_Lipschitz}, and $\sigma(y)$ is symmetric positive definite for all $y\in \R^d$.
		\item $\sigma$ is differentiable everywhere and for all $y\in \R^d$.
		\begin{equation}\label{eq:no_curl_cond_}
			\frac{\partial\sigma(y)}{\partial y_k} \sigma(y)^{-1} e_j  = \frac{\partial\sigma(y)}{\partial y_j} \sigma(y)^{-1} e_k,
		\end{equation}
		where $e_j$ is the $j$-th canonical basis vector of $\R^d$.
	\end{enumerate}
	Then there exists a solution $(\tilde f,\tilde \mu, \tilde \sigma)$ to Problem \ref{prob:identifiability} such that $\tilde \sigma$ is isotropic.
\end{theorem}
\begin{proof}
Using Proposition \ref{prop:1}, it suffices to find an invertible function $g$ such that $g(Z_t)$ is governed by an SDE with an isotropic diffusion coefficient. An SDE for which such a function exists is called reducible and \textit{\ref{thm:canonical_sigma_}.i)} and \textit{(\ref{thm:canonical_sigma_}.ii)} are necessary and sufficient conditions for an SDE to have this property. See \cite[Proposition 1]{ait2008closed} for a formal statement of that result and respective proof. The proof of Theorem \ref{thm:canonical_sigma_} then follows by letting $\tilde f = f \circ g^{-1}$ and defining $\tilde \mu$ in terms of $g$, $\mu$ and $\sigma$, using It\^o's Lemma \cite{kunita1967square}.
\end{proof}

Finally, we provide a lemma to further the understanding of Theorem \ref{thm:canonical_sigma_}, in particular when \textit{(\ref{thm:canonical_sigma_}.ii)} holds.

\begin{lemma}\label{lemma:whatcansigmabe}
Suppose that $\sigma$ satisfies \textit{(\ref{thm:canonical_sigma_}.i)}, then any of the following conditions are sufficient for \textit{(\ref{thm:canonical_sigma_}.ii)} to hold.
\begin{enumerate}[(\ref{lemma:whatcansigmabe}.i)]
    \item The latent dimension is $1$ ($d=1$) and $\sigma(y)$ is positive.
    \item $\sigma(y)$ is a positive diagonal matrix and the $i$-th diagonal element depends only on coordinate $i$, that is, there exist functions $f_i:\R\to \R$ such that $\sigma(y)_{ii} = f_i(y_i)$, for all $y\in\R^d$.
    \item There exists a $d\times d$ invertible  matrix $M$ and a function $\varLambda:\R^d\to \R^{d\times d}$ such that $\sigma(y) = M \varLambda(M^{-1} y) M^T$, and $\varLambda(y)$ satisfies \textit{(\ref{lemma:whatcansigmabe}.ii)}, for all $y\in\R^d$.
    \item \label{item:hessian_convex} $\sigma^{-1}$ is the Hessian of a convex function.
\end{enumerate}
Moreover, (\ref{lemma:whatcansigmabe}.\ref{item:hessian_convex}) is also necessary. 
As an example, a Brownian motion is a reducible SDE. This condition also holds in many cases of practical interest, see examples in \cite{ait2002maximum}. 
\end{lemma}

\begin{proof}
See Appendix~\ref{sec:sigma_lemma_proof}
\end{proof}

\section{Estimating the Latent SDE using a VAE}\label{sec:VAE_def}

Motivated by the previous section, in this section we assume the latent space is governed by an SDE with an an isotropic diffusion coefficient, and describe  a method for recovering $(f, \mu)$, using a VAE. 
While $\{X_t\}_{t \in \T}$ is a stochastic process defined for all $t \in \T$, in practice, we sample $X_t$ at discrete times and, for ease of exposition, we assume unless stated otherwise that the sampling frequency is constant. 

\subsection{Generative Model}
 
\begin{figure}
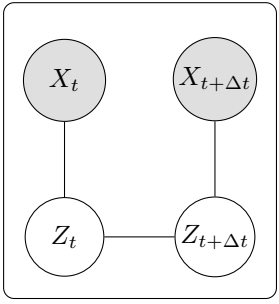

\centering
  \tikz{
 \node[latent,fill, xshift=2cm] (zt) {$Z_{t+\Delta t}$}; %
 \node[latent] (z) {\phantom{a}\,\,$Z_t$\,\,\phantom{a}}; %
  \node[obs, above=of zt] (xt) {$X_{t+\Delta t}$};%
 \node[obs,above=of z] (x) {\phantom{a}\,\,$X_t$\,\,\phantom{a}}; %
 \plate [inner sep=.25cm,yshift=.2cm] {plate1} {(x)(xt)(zt)(z)} {}; %
 \draw (xt) -- (zt);
 \draw (z) -- (zt);
 \draw (x) -- (z);
 }
 \caption{Probabilistic graphical model for our generative model. The ambient variables $X_t$ and $X_{t + \deltat}$ only depend on each other through the latent variables $Z_t$ and $Z_{t + \deltat}$.}
 \label{fig:PGM}
\end{figure}

In order to learn the decoder and the drift coefficient, we consider pairwise consecutive time series observations $\bX =\nobreak (X_{t+\deltat}, X_t)$, which correspond to the latent variables $\bZ = (Z_{t+\deltat}, Z_t)$. Accordingly, we consider the following conditional generation model, with model parameters ${\bm{\phi}} = (f, \mu, \gamma)$.
\begin{equation}\label{eq:Xt_factorization}
\begin{split}
p_{\bm{\phi}}(\bX\! &, \bZ) = \\ & p_f(X_{t+\deltat}|Z_{t+\deltat}) p_{\mu}(Z_{t+\deltat}|Z_t) p_f(X_t| Z_t) p_\gamma(Z_t).
\end{split}
\end{equation}
where
\begin{itemize}
\item The terms $p_f(X_{t+\deltat}|Z_{t+\deltat})$ and $p_f(X_t|Z_t)$ are defined by \eqref{eq:fplusnoise}, which implies that
\begin{equation}\label{eq:Xt_epsilon}
p_f(X_t|Z_t) = p_\epsilon(X_t - f(Z_t)),	
\end{equation}
where $p_\epsilon$ is the probability distribution function of $\epsilon_t$.
\item The prior distribution on the latent space is given by $p_\gamma(Z_t)$. 
This term is added to ease the training of the VAE.
\item The term $p_{\mu}(Z_{t+\deltat}|Z_t)$ is related to the SDE dynamics. 
In order to model this equation with a conditional generation model, we use the Euler-Maruyama method, which provides an approximation for the distribution of $Z_{t+\Delta t}$ that is valid if $\Delta t$ is small enough. 
Recalling that $Z_t$ is a solution to an SDE with an isotropic diffusion coefficient, and drift coefficient $\mu:\R^{d}\to \R^d$, we have
\begin{equation}\label{eq:Zt_SDE}
Z_{t+\deltat} \approx Z_t + \mu(Z_t) \deltat + W_{t+\deltat}-W_{t}.
\end{equation}
Since $W_{t+\deltat}-W_{t}$ is distributed as a multivariate centered Gaussian variable with variance $\deltat I_d$, we define
\begin{equation}\label{eq:Zt_pSDE}
\begin{split}
p_\mu &(Z_{t+\deltat}|Z_t) = \\ & \frac1{\sqrt{2\pi \deltat}^d}\exp\left(-\frac{\|Z_{t+\deltat} - Z_t - \mu(Z_t) \deltat\|^2}{2 \deltat}\right).
\end{split}
\end{equation}
\end{itemize}
For the probabilistic generative model we consider, the ambient variables $X_t$ and $X_{t + \deltat}$ only depend on each other through the latent variables $Z_t$ and $Z_{t + \deltat}$. The corresponding Markov network model is drawn in Fig.~\ref{fig:PGM}.

\subsection{VAE encoder and training loss}
We describe an encoder $q_{\bm{\psi}}(\bZ| \bX)$, that approximates the true posterior $p_{{\bm{\phi}}}(\bZ| \bX)$, which is computationally intractable. It follows from \eqref{eq:Xt_factorization} and Fig.~\ref{fig:PGM} that $Z_t$ is independent of $X_{t+\deltat}$ conditioned on $Z_{t+\deltat}$, and $Z_{t+\deltat}$ is independent of $X_{t}$ conditioned on $Z_t$, thus we can factorize
\begin{equation}\label{eq:p_factorization}
p_{{\bm{\phi}}}(\bZ| \bX) = p_{{\bm{\phi}}}(Z_{t+\deltat}| X_{t+\deltat}, Z_t) p_{{\bm{\phi}}}(Z_t| X_t).
\end{equation}
Accordingly, we can factorize our encoder as \begin{equation}\label{eq:encoder_decomposition}
q_{\bm{\psi}}(\bZ| \bX) = q_{\bm{\psi}_1}(Z_{t+\deltat}|X_{t+\deltat}, Z_t) q_{\bm{\psi}_2}(Z_{t}|X_{t}).
\end{equation}
Regarding the training loss, let $\mathcal{D}=\{x_{t+\deltat}, x_{t}\}_{t\in \bm{T}}$ be the observed data, already paired into consecutive observations, and $q_{\mathcal{D}}$ the empirical distribution in $\mathcal{D}$. We then train the VAE by minimizing the loss
\begin{align}
& \nonumber \mathcal{L}({\bm{\phi}},{\bm{\psi}})\\
\nonumber &\hspace{5pt}= D_{KL}\left(q_{\bm{\psi}}(\bZ| \bX)q_{\mathcal{D}}(\bX)\;\big{\|}\; p_{\bm{\phi}}(\bZ| \bX)q_{\mathcal{D}}(\bX) \right) \\ &\hspace{130pt}- \E_{q_{\mathcal{D}}(\bX)}\left[p_{{\bm{\phi}}}(\bX)\right],
\label{eq:L_def_KL} \\ 
&\hspace{5pt}= \E_{q_{\mathcal{D}}(\bX)}\left[\E_{q_{\bm{\psi}}(\bZ| \bX)}
\left[\log q_{\bm{\psi}}(\bZ| \bX) -\log p_{{\bm{\phi}}}(\bX\!, \bZ) \right]\right]. 
\label{eq:L_def}
\end{align}

Here \eqref{eq:L_def_KL} can be thought of as the negative evidence lower bound; minimizing it forces $q_{\bm{\psi}}(\bZ| \bX)$ to approximate $p_{\bm{\phi}}(\bZ| \bX)$ while maximizing the likelihood of $p_{\bm{\phi}}(\bX)$ under the distribution $q_{\mathcal{D}}$.

To calculate \eqref{eq:L_def}, we use the reparametrization trick \cite{kingma2013auto} to backpropagate through the SDE, and the exact form of the KL divergence between two Gaussians which we describe in Appendix~\ref{sec:VAE_loss}. The training algorithm then proceeds as a regular VAE.

\subsection{An approximate encoder}

For training the VAE, it is convenient to consider the simplified encoder:
\begin{equation}\label{eq:encoder_decomposition_approximate}
\tilde q_{\bm{\psi}}(\bZ| \bX) = \tilde q_{\bm{\psi}}(Z_{t+\deltat}|X_{t+\deltat}) \tilde q_{\bm{\psi}}(Z_{t}|X_{t}).
\end{equation}
This decomposition allows for using the same encoder twice, and therefore eases the training of the VAE. If $\epsilon=0$ in \eqref{eq:Xt_epsilon}, and since $f$ injective, we would be able to determine $Z_{t+\deltat}$ from $X_{t+\deltat}$. In particular, that would imply $Z_{t+\deltat}$ was conditionally independent of $Z_t$, given $X_{t+\deltat}$, and that \eqref{eq:encoder_decomposition_approximate} was exact. Although the noise $\epsilon$ is not $0$, we assume it is relatively small compared with the noise related to the SDE term $p_{\mu}(Z_{t+\deltat}| Z_t)$. Intuitively, that implies $X_{t+\deltat}$ gives much more information about $Z_{t+\deltat}$ than $Z_t$, and we can consider the approximation
\begin{equation}\label{eq:p_approx_factorization}
p_{{\bm{\phi}}}(Z_{t+\deltat}| X_{t+\deltat}, Z_t) \approx p_{{\bm{\phi}}}(Z_{t+\deltat}| X_{t+\deltat}),
\end{equation}
without losing too much information. We formalize this argument in the following proposition, using mutual information.
On one hand, the quantity $I(Z_{t+\deltat}; X_{t+\deltat})$, measures the information one gets of $Z_{t+\deltat}$ by learning $X_{t+\deltat}$, and $I(Z_{t+\deltat}; Z_t|X_{t+\deltat})$ measures the additional information one gets of $Z_{t+\deltat}$ by further knowing $Z_t$.
On other hand, the KL divergence term that appears in the definition of mutual information will be the same that appears in the training loss of the VAE \eqref{eq:L_def}. Our assumption that the noise $\epsilon$ is small compared to the SDE term can be formalized as $I(Z_{t+\deltat}; X_{t+\deltat}) \gg I(Z_{t+\deltat}; Z_t)$, and this hypothesis can be used to justify our argument.

\begin{proposition}\label{proposition:encoder_decomp}
If $I(Z_{t+\deltat}; Z_t) \ll I(Z_{t+\deltat}; X_{t+\deltat})$, then
\begin{equation}
I(Z_{t+\deltat}; Z_t|X_{t+\deltat})  \ll I(Z_{t+\Delta t}; X_{t+\Delta t}),
\end{equation}
and $I(Z_{t+\deltat}; X_{t+\deltat}, Z_t)\approx I(Z_{t+\deltat}; X_{t+\deltat})$.
\end{proposition}
\begin{proof}
The proof follows from applying the chain rule and non-negativity of the mutual information. We present the details and recall the definition of mutual information in Appendix~\ref{sec:encoder_proof}.
\end{proof}

\section{Identifiability}\label{sec:identifiability}

In this section, we return to the topic of identifiability. Previously, we showed that we can assume that the diffusion coefficient is isotropic, and introduced the prior parameter $\gamma$ as a mechanism to ease the training of the VAE. Here we provide identifiability results for the remaining model parameters ${\bm{\phi}} = (f, \mu, \gamma)$.

A crucial element of our analysis concerns the probability distribution of $\bX = (\hat X_{t+\deltat}, \hat X_t)$, that is, the distribution of pairwise consecutive data points. The probability distribution of $\bX$ is defined by ${\bm{\phi}}$, through equation \eqref{eq:Xt_factorization}, by integrating over $\bZ$. Suppose that ${\bm{\phi}^*} = (f^*, \mu^*, \gamma^*)$ are the true model parameters of the data, and ${\bm{\phi}} = (f, \mu, \gamma)$ are model parameters such that
\begin{equation}\label{eq:generative_coincide}
p_{{\bm{\phi}}}(x_{t+\deltat}, x_t) = p_{{\bm{\phi}}^*}(x_{t+\deltat}, x_t)\quad \forall x_{t+\deltat}, x_t \in \R^n,
\end{equation}
Then, since these two generative models coincide, and the latent space is unknown, it is not possible to determine which of these two models provides a description, through equation \eqref{eq:Xt_factorization}, of the true latent space. In other words, both models provide plausible descriptions of the latent space. 

It is therefore important, for identifiability purposes, to characterize all parameter configurations where the generative models coincide.
In the following theorem, we show that if the generative models coincide, then the corresponding model parameters are equal up to an isometry. Recalling Proposition \ref{prop:1}, it becomes clear that it is only possible to recover ${\bm{\phi}}$ up to an isometry: if $g$ is an isometry and $Z_t$ is a solution to an SDE with an isotropic diffusion coeffient, then $g(Z_t)$ is a solution to another SDE also with an isotropic diffusion coeffient.

\begin{theorem}\label{thm:identifiability_main}
Suppose that the true generative model of $\bX$ has parameters ${\bm{\phi}}^*= (f^*, \mu^*, \gamma^*)$, and that the following technical conditions hold:
\begin{enumerate}
	\item The set $\{x\in \mathcal X|{\bm{\varphi}}_\epsilon(x)=0\}$ has measure zero, where ${\bm{\varphi}}_\epsilon$ is the characteristic function of the density $p_\epsilon$ defined in \eqref{eq:Xt_epsilon}.
	\item $f^*$ is injective and differentiable.
	\item $\mu^*$ is differentiable almost everywhere.
\end{enumerate}
Then, for almost all values of $\deltat$,\footnote{Specifically, there is a finite set $S$ such that if $\deltat\notin S$, the condition holds.} if ${\bm{\phi}} = (f, \mu, \gamma)$ are other model parameters such that \eqref{eq:generative_coincide} holds, 
then ${\bm{\phi}}$ and ${\bm{\phi}}^*$ are equal up to an isometry. That is, there exists an orthogonal matrix $Q$ and a vector $b$, such that for all $z\in \R^d$:
\begin{equation}
f(z) = f^*(Q z + b),
\end{equation}
\begin{equation}
\mu(z) = Q^T \mu^*(Q z + b),
\end{equation}
and
\begin{equation}\label{eq:ident_p_gamma}
p_{\gamma}(z) = p_{\gamma^*}(Q z + b).
\end{equation}
\end{theorem}

The proof of Theorem \ref{thm:identifiability_main} is closely related with the theory developed in \cite{khemakhem2019variational}, and is available in Appendix~\ref{sec:proof_thm3}. Finally, we show the VAE framework presented in this paper can obtain the true model parameters in the limit of infinite data.
\begin{theorem}\label{thm:VAE_convergence}
Let $\{q_{{\bm{\psi}}}(\bZ| \bX)\}_{\bm{\psi}\in \bm{\Psi}}$ be an encoder that can be factorized as in \eqref{eq:encoder_decomposition}, where $\bm{\Psi}$ includes all parameter configurations of the encoder, and assume the following:
\begin{itemize}
    \item The family $\{q_{{\bm{\psi}}}(\bZ| \bX)\}_{\bm{\psi}\in \bm{\Psi}}$ includes $p_{{\bm{\phi}}^*}(\bZ| \bX)$, 
    \item $\mathcal{L}({\bm{\phi}},{\bm{\psi}})$ is minimized with respect to both $\bm{\phi}$ and $\bm{\psi}$.
\end{itemize}
Then, in the limit of infinite data, we obtain the true model parameters ${\bm{\phi}}^*=(f^*, \mu^*, \gamma^*)$, up to an isometry.
\end{theorem}

\begin{proof}
See \cite[Supplemental Material B.6]{khemakhem2019variational}.
\end{proof}

We note that in this result we consider a general encoder that can be factorized as in \eqref{eq:encoder_decomposition}, and not the simpler encoder that we introduce in  \eqref{eq:encoder_decomposition_approximate}. Empirically, we observe that using this simplification introduces a model generalization error that is small compared with the data generalization error.


\section{Practical considerations}\label{sec:practical_considerations}
We made number of simplifying assumptions that may not hold in practical cases.
Here we discuss some of their implications on the proposed method. 

\subsection{Variable sampling frequency}

In order to simplify the exposition of the results, we have assumed that the sampling frequency is fixed. However the proposed framework can also accommodate variable sampling frequency with some modifications. Specifically, for two consecutive observations at times $t_1$ and $t_2$, \eqref{eq:Zt_SDE} becomes
\begin{equation}\label{eq:Zt_SDE_t}
	Z_{t_2} \approx Z_{t_1} + \mu(Z_{t_1}, t_1) (t_2-t_1) + W_{t_2}-W_{t_1},
\end{equation}
and $p_\mu (Z_{t_2}|Z_{t_1})$ is defined analogously to \eqref{eq:Zt_pSDE}. Furthermore, Theorems \ref{thm:identifiability_main} and \ref{thm:VAE_convergence} also hold for this modification. 

We note however that this approach depends on the validity of approximation \eqref{eq:Zt_SDE_t}. If $t_2 - t_1$ is too large, an adjustment of the underlying integrator may be necessary. One possible integrator is to split the interval in multiple sub-intervals, use Euler-Maruyama in each sub-interval, and use the parametrization trick for training. Other possible integrators use diffusion bridges or a multi-resolution MCMC approach inspired by the results in \cite{roberts2001inference, kou2012multiresolution}.
\subsection{SDEs with time dependence}

While our primary focus is on time-independent SDEs due to their prevalence in the literature, we additionally describe how our method can also be used for time-dependent SDEs. 
Time-dependent SDEs have relevant applications in finance, see for example \cite{hull1990pricing, black1990one}.
We consider a similar conditional generation model as in \eqref{eq:Xt_factorization}, where
\eqref{eq:Zt_SDE} should be rewritten as \eqref{eq:Zt_SDE_t}, which implies that \eqref{eq:Zt_pSDE} becomes
\begin{align*}
 p_\mu (Z_{t_2}|& Z_{t_1}, t_1) = \frac1{\sqrt{2\pi (t_2-t_1)}^d} \\ & \exp\left(-\frac{\|Z_{t_2} - Z_{t_1} - \mu(Z_{t_1}, t_1) (t_2-t_1)\|^2}{2 (t_2-t_1)}\right).
\end{align*}
The encoder can also depend on time by appending the time value to the last linear layer of the encoder, if the approximation given by \eqref{eq:encoder_decomposition} is insufficient. 
Modifying Theorems~\ref{thm:identifiability_main} and \ref{thm:VAE_convergence} to accommodate time-dependent drift coefficients is straightforward, see Theorem~\ref{thm:identifiability_arbitrary} for an example on how Theorem~\ref{thm:identifiability_main} is also valid for time-dependent SDEs.
For Theorem~\ref{thm:canonical_sigma_}, the crucial part is the following extension of \cite[Proposition 1]{ait2008closed} to time-dependent SDEs, which we prove in Appendix~\ref{sec:supplementary}.A.

\begin{theorem}[Multivariate time-dependent Lamperti transform]
	Suppose that $\{Y_t\}_{t\in \T}\in \R^d$ is a solution to the SDE:
	\begin{equation}\label{eq:general_Y_t}
		dY_t = \mu(Y_t, t)\d t + \sigma(Y_t, t)\d W_t,
	\end{equation}
	where $\mu:\R^d\times\T \to \R^d$ and $\sigma:\R^d\times\T\to \R^{d\times d}$.
	Moreover, suppose the following conditions are satisfied:
	\begin{enumerate}[(i)]
		\item $\mu$ and $\sigma$ are globally Lipschitz, that is, \eqref{eq:mu_sigma_Lipschitz} holds, and $\sigma(y, t)$ is symmetric positive definite for all $y\in \R^d,t\in \T$.
		\item $\sigma$ is differentiable everywhere and for all $y\in \R^d$,  $t\in \T$ and $j,k \in \{1,\dots,d\}$
		\begin{equation}\label{eq:no_curl_cond}
			\frac{\partial\sigma(y, t)}{\partial y_k} \sigma(y, t)^{-1} e_j  = \frac{\partial\sigma(y, t)}{\partial y_j} \sigma(y, t)^{-1} e_k,
		\end{equation}
		where $e_j$ is the $j$-th canonical basis vector of $\R^d$.
	\end{enumerate}
	Then there exists a function $g:\R^{d}\times\T \to \R^d$ and $\tilde \mu:\R^{d}\times\T\to \R^d$ such that $Y_t = g(Z_t, t)$ and $\{Z_t\}_{t\in \T}$ is a solution to the SDE:
	\begin{equation}\label{eq:Z_t_canonical_}
		dZ_t = \tilde \mu(Z_t, t)\d t + \d W_t,
	\end{equation}\label{thm:time_lamperti}
\end{theorem}
\vspace{-20pt}
Using the time-dependent Lamperti transform combined with the time dependence results in Theorem~\ref{thm:identifiability_arbitrary} allows for the straightforward extension to time dependent SDEs. 

\subsection{Determining the latent dimension}\label{sec:latent_dimension}

In order to learn the latent dimension, we suggest using the following architecture search heuristic. Instead of considering an isotropic diffusion coefficient, we set $\sigma(y)= D$ for all $y\in \R^d$, where $D$ is a diagonal matrix with learnable diagonal entries. Starting with a guess for the latent dimension, we increase it if the image reconstruction is unsatisfactory, and decrease it if some of the diagonal entries of $D$ are close to $0$ (adding an $\ell_1$ regularization to the diagonal entries of $D$ will promote sparsity and help drive some of its values to $0$). 

\emph{Using the likelihood in the linear case.}
As a first step in obtaining theoretical guarantees for determining the latent dimension, we provide a result on using the likelihood to recover the true latent dimension for the case where $f, {\bm{\psi}}$ are linear functions. 
\begin{theorem}[Latent size recovery with the likelihood] \label{thm:latent_size}
    Suppose the true generative model of $\mathbf{X}$ is generated according to a full rank \emph{linear} transformation of a latent SDE $Z_t$ with conditions on $\mu, \sigma$ as above
    \begin{align*}
        X_t &= AZ_t, \quad A \in \mathbb{R}^{n \times d} \\
        dZ_t &= \mu(Z_t, t) dt + \sigma(Z_t, t) dW_t, \quad Z_t \in \mathbb{R}^d.
    \end{align*}
    Moreover suppose that $\rank(A) = d$ and let the estimate of $A$ with dimension $j$ be $\hat{A}_j$.
    Then in the limit of infinite data, the model with latent size $j$ satisfying
    \begin{equation}
    \max_{j \in \mathbb{N}} \log \mathcal{L}(\hat{A}_j), \quad \hat{A}_j \in \mathbb{R}^{n \times j}
    \label{eq:aic}
    \end{equation}
    will recover the proper latent dimension $j = d$. 
\end{theorem}
\begin{proof}
The proof involves considering the likelihood of the transformed variables for different latent dimensions.
See Appendix~\ref{sec:latent_proof}.C for more details.
\end{proof}

\emph{Estimating the diffusion coefficient.}
In Appendix~\ref{sec:supplementary}.B, we present an interpretability result that considers learnable diffusion coefficients. Unfortunately, this result requires conditions that do not apply for simpler SDEs, such as Brownian random walks, therefore we decided to present Theorem \ref{thm:identifiability_main} in the paper instead.
\section{Experiments}\label{sec:results}
We consider 4 synthetic and one real-world datasets to illustrate the efficacy of SDE-VAE.

\subsection{Datasets}
\subsubsection{Moving Yellow Ball}
For this dataset, we simulate the stochastic motion of a yellow ball moving according to a given SDE using the Euler-Maruyama method. That is to say, the $x$ and $y$ coordinates of the center of the ball are governed by an SDE with an isotropic diffusion coefficient, and the drift coefficients are defined for $(x,y)\in \R^2$ as follows.
\begin{align*}
	\text{\textbf{Constant}:}\quad & \mu(x,y) = (-1/4,1/4);\\
	\text{\textbf{OU}:}\quad & \mu(x,y) = (-4x,-4y); \\
	\text{\textbf{Circle}:}\quad & \mu(x,y) = (-x-3y,y-3x);
\end{align*}
where \textbf{OU} stands for the Ornstein-Uhlenbeck process. 
The latent space dimension is $2$, corresponding to the $x$ and $y$ coordinates of the ball.
Fig.~\ref{fig:12_balls} shows an example of the movement of the balls. 
We train the model on one realization of the SDE for 1000 time steps with $\Delta t = 0.01$. We rescale the realization of the SDE so that the $x$ and $y$ coordinates of the ball are always between $0$ and $1$; in practice, this only changes the map from latent to ambient space, and should not affect the ability of our method to recover the latent SDE realization.
We add an extension to this dataset using \textit{5 Moving Blue Balls} with a 10 dimensional latent space. 
We study an \textbf{OU} process where each of the balls reverts to a specific section within the image. 
This dataset is challenging because of the number of objects and due to the changes in illumination when balls overlap. 
\subsubsection{Moving Red Digits}
To further investigate the generative properties of the proposed method, we consider images of 2 digits from the MNIST dataset moving according to an SDE in the image plane, similarly to the dataset above, and use the Euler-Maruyama method to simulate the spatial positions of the two digits. In this case, the latent space is 4-dimensional, corresponding to the $x$ and $y$ coordinates of the center of each of the digits. Letting $\bm{x}\in \R^2$ and $\bm{y}\in \R^2$ be the $x$ and $y$ coordinates of each ball, then the drift coefficient is defined for $(\bm{x},\bm{y})\in \R^4$ as follows.
\begin{align*}
	\text{\textbf{Constant}:}\quad & \mu(\bm{x},\bm{y}) = (\bm{0},\bm{0});\\
	\text{\textbf{OU}:}\quad & \mu(\bm{x},\bm{y}) = (-\bm{x}-1,-\bm{y}+1); \\
	\text{\textbf{Circle}:}\quad & \mu(\bm{x},\bm{y}) = (-x_0-2x_1,-x_1+2x_0, \\ & \hspace{50pt} -y_0-2y_1, -y_1+2y_0); 
\end{align*}
Here, $\mu(\bm{x},\bm{y})\in \R^4$ and the product of vectors is defined entry-wise. For each digit pair, we generate 10 trajectories of the SDE with 100 frames in each trajectory and $\Delta t = 0.01$. In this dataset, we also rescale the realization of the SDE so that the $x$ and $y$ coordinates of the digits are always between $0$ and $1$.
All diffusion matrices are the identity matrix. 

\subsubsection{Wasserstein Interpolation}
\begin{figure}[ht!]
	\centering
	\includegraphics[width=\textwidth]{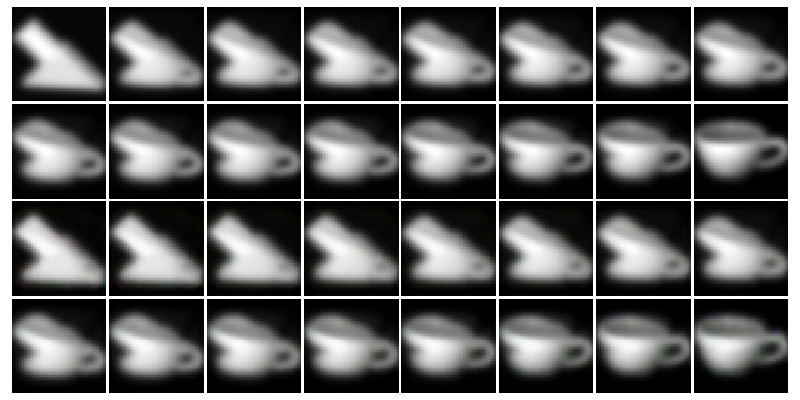}
	\caption{Sampling a Brownian bridge in the Wasserstein OU dataset. Top two rows: ground truth generated bridge between endpoints. Bottom two rows: sampled bridge between endpoints from latent space using proposed method. First, the endpoint images are embedded into the latent space. Then, we sample a Brownian bridge between the two latent points. Finally, we decode the latent bridge using the decoder.}
	\label{fig:wass_ex}
\end{figure}

We are interested in investigating the algorithm's efficacy on a non movement dataset by generating a series of interpolations between two images according to their Wasserstein barycenters \cite{solomon2015convolutional}. 
This experiment aims to consider the method's performance on a more complicated dataset than the previous two. 
For instance, spatial position-based encodings of the objects in the images will not work in this case. 
We sample images from the COIL-20 dataset \cite{nene1996columbia} to interpolate according to the realization of an SDE (an example is given in the third row of Fig.~\ref{fig:reconstruction}).
To further illustrate the idea, Fig.~\ref{fig:wass_ex} shows the output of the decoder when interpolating between two images using a Brownian bridge.
We simulate a 1-D SDE which determines the relative weight of each image within the interpolation. 
The drift coefficient is defined as follows.
\begin{align*}
	\text{\textbf{OU}:}\quad & \mu(x) = -2x\quad \forall x\in \R;  \hspace{50pt} \\
	\text{\textbf{Double Well}:} \quad & \mu(x) = 2x(1-x^2) \quad \forall x\in \R; \\
    \text{\textbf{GBM}:}\quad & \mu(x) = \frac12 x; \:\: \sigma(x) = x;
\end{align*}
where \textbf{GBM} refers to geometric Brownian motion.
We simulate 1000 images with $\Delta t = 0.01$. 
The realization of the SDE is rescaled so that the relative weight of each image is always between $0$ and $1$.
For the \textbf{GBM} case, the Lamperti transform of the original SDE results in an It\^o process with 0 drift and unit diffusion.
We therefore compare the learned latent $\mu$ to the constant zero drift.
All diffusion coefficients are constant unit apart from the GBM case.

\subsubsection{Moving Ball with Wasserstein Background}
As an additional challenge, we will consider a dataset where we have a 3 dimensional SDE where one component modulates the background while another moves a ball in the foreground. 
We use the Wasserstein interpolation as the background process and the yellow ball as the foreground. 
This is a complicated dataset that adds occlusion and multiple moving parts to the underlying SDE. 
We choose the following two drift coefficients, one OU process similar to previous experiments and another process similar to the Cox-Igersoll-Ross process with anisotropic diffusion. Letting $x, y$ be the $x,y$ coordinates of the ball, and $z$ be the Wasserstein barycenter, we define the drift coefficients, and diffusion coefficient for the anisotropic SDE, as follows.
\begin{align*}
	\text{\textbf{OU}:}\quad & \mu(x, y, z) = (-x, -2y, -3z); \\
	\text{\textbf{Cauchy}:} \quad & \mu(x, y, z) = \left (\frac{-x}{1+x^2}, \frac{-2y}{1+y^2}, \frac{-3z}{1+z^2} \right )  \\
	\text{\textbf{Anisotropic}:}\quad & \mu(x, y, z) = (-x, -2y, 0.6-0.3z); \\
	& \sigma(x, y, z) = \begin{pmatrix}
	1 & 2 & 0 \\
	2.5 & 3 & 0 \\
	0 & 0 & \sqrt{z}
	\end{pmatrix}
\end{align*}
Taking the multivariate Lamperti transform of the anisotropic case, we obtain a new drift of the form 
$$\mu(x,y,z) = \left(1.5 x - 2y, -1.25x + y, \tfrac{1.4 - 1.2 z}{4\sqrt{z}}\right)$$
with which we estimate the recovery efficacy. 

\subsubsection{Fluorescent DNA}
\label{sec:dna}
The last dataset consists of videos of a DNA molecule floating in solution undergoing random thermodynamic fluctuations as described in the work of \cite{lameh2020controlled}. The videos undergo minimal pre-processing, through histogram equalization based on \cite{zuiderveld1994contrast} and normalization of the pixel values. The ``ground truth'' latent variables are obtained by segmenting the molecule using a method similar to the one described in \cite{lameh2020controlled} and using the center of the segmented molecule. 
Using these latent variables as ground truth, we compute the best affine mapping between these and the ones estimated using our method and report the results in Table~\ref{tab:sde_results}.  
For this experiment we analyze three datasets, two with additional noise added in one direction of the molecule and the other with no noise added (denoted by $V=1,2$ and $V=0$ respectively).
The $V=1$, $V=2$ datasets have an anisotropic diffusion coefficient, with additional intensity given in the $y$ variable whereas in the $V=0$ case the noise is isotropic with the identity matrix as the diffusion following equations (3) and (5) from \cite{lameh2020controlled}.
In both cases, the ground truth should be a random walk corresponding with no drift, that is, $\mu = (0, 0)$. We use these as the ground truth drift values and compute the MSE between the estimated and the theoretical drift. Since for this dataset the diffusion coefficient is not known, we compute the latent variables using an affine transformation, rather than an orthogonal transformation. 
Since the datasets are fixed, we repeat the experiment 5 times with different initializations of the neural networks to report the values in the table. 
{
\renewcommand\arraystretch{1.2}
\begin{table*}
	\newcommand{\timesten}{\text{\footnotesize $\times10$}}
	\centering
	\footnotesize
	\begin{tabular}{@{}lccccc@{}} \toprule
		Dataset & SDE Type & $\mathcal{L}_\text{latent}$ & $\mathcal{L}_\mu$  & $\mathcal{L}_\mu$ CRLB  & Reconstruction MSE\\ \midrule 
		
		\multirow{3}{*}{Balls} & Constant &  $1.08 (\pm 0.47) \timesten^{0\phantom{-}}$ & $5.07 (\pm  2.99) \timesten^{-1} $ & 	\multirow{3}{*}{$2.00 \timesten^{-1}$} & $1.71 (\pm 1.85) \timesten^{-2}$ \\ 
		& OU  &  $1.41(\pm 1.20) \timesten^{-2}$ & $9.20 (\pm 2.23) \timesten^{-1}$ & & $3.23 (\pm 12.0) \timesten^{-3}$ \\ 
		
		& Circle & $4.65(\pm 2.56) \timesten^{-2}$ & $1.47 (\pm 0.62) \timesten^{0\phantom{-}}$ &  & $5.10 (\pm 3.55) \timesten^{-3}$ \\  
		Multiple Balls & OU & $2.90(\pm0.79)\timesten^{-1}$ & $2.61(\pm 0.51) \timesten^{0}$ & $1.00 \timesten^{0\phantom{-}}$ & $1.69(\pm1.05)\timesten^{-2}$\\

		\cline{1-6}

		\multirow{3}{*}{Digits} & Constant & $4.36(\pm 1.11) \timesten^{-1}$ & $7.74(\pm 4.65) \timesten^{-1}$ &  	\multirow{3}{*}{$4.00 \timesten^{-1}$} & $4.58 (\pm 1.13) \timesten^{-3} $ \\ 
		& OU &  $1.99(\pm 0.47) \timesten^{-1}$ & $7.69(\pm 4.91) \timesten^{-1}$ & & $2.57 (\pm 0.96) \timesten^{-3} $ \\ 
		& Circle & $2.38 (\pm 1.15) \timesten^{-1} $ & $2.15 (\pm 0.48) \timesten^{0\phantom{-}}$ & & $3.02 (\pm 0.88) \timesten^{-3} $ \\ \cline{1-6}

		\multirow{3}{*}{Wasserstein} & OU & $3.53(\pm 4.31) \timesten^{-2}$  & $8.79(\pm 1.21) \timesten^{-1}$ & \multirow{3}{*}{$1.00 \timesten^{-1}$} & $5.55(\pm 3.03) \timesten^{-4}$ \\ 
		& Double Well & $5.20(\pm 6.40) \timesten^{-2}$ & $1.45(\pm 0.76) \timesten^{0\phantom{-}}$ &  & $2.08(\pm 3.33)\timesten^{-3}$ \\ 
		& GBM & $3.28(\pm 2.32) \timesten^{-1}$ & $1.23(\pm 2.45) \timesten^{-1}$  &  & $ 2.23 (\pm 2.81) \timesten^{-3} $ \\ \cline{1-6}
		
		\multirow{3}{*}{Ball + Wasserstein} & OU & $4.23(\pm 2.70) \timesten^{-2}$  & $7.08(\pm 2.79) \timesten^{-1}$ & \multirow{3}{*}{$3.00 \timesten^{-1}$}& $1.76 (\pm 0.68) \timesten^{-2}$  \\ 
		& Cauchy & $1.24(\pm 0.90) \timesten^{-1}$ & $8.82(\pm 7.36) \timesten^{-1}$ & & $1.91(\pm 0.81)\timesten^{-2}$\\ 
		& Anisotropic & $1.03(\pm0.82)\timesten^{0\phantom{-}}$ &$1.74(\pm0.50)\timesten^{0\phantom{-}}$ & & $1.90(\pm0.43)\timesten^{-2}$ \\ 
		\bottomrule
		Ball + $\mathcal{N}(0,4)$ & \multirow{6}{*}{OU} & $1.16(\pm 0.51)\timesten^{-2}$ & $1.22 (\pm 0.39) \timesten^{0\phantom{-}}$ &\multirow{6}{*}{N/A} & $8.99(\pm 5.28) \timesten^{-3}$\\
		
		Ball + $t$-noise & & $1.14(\pm 0.30) \timesten^{-2}$ &$1.29 (\pm 0.17) \timesten^{0\phantom{-}}$ & & $6.99 (\pm 1.99) \timesten^{-3}$ \\
		Wasserstein + $\mathcal{N}(0,4)$ & & $1.16(\pm 0.51) \timesten^{-2}$ & $9.51 (\pm 5.63)\timesten^{-1}$ & & $2.09 (\pm 0.43) \timesten^{-3}$ \\
		Wasserstein + $t$-noise & & $3.83(\pm 0.55)\timesten^{-2}$& $9.95(\pm 1.12)\timesten^{-1}$ & & $1.61 (\pm 0.43)\timesten^{-3}$ \\
		Ball + Wasserstein + $\mathcal{N}(0,4)$ & & $8.57 ( \pm 4.33 ) \timesten^{-2}$ & $1.34(\pm 1.05)\timesten^{0\phantom{-}}$ & & $2.23 (\pm 0.24) \timesten^{-2}$ \\
		Ball + Wasserstein + $t$-noise & & $9.62 (\pm 5.22) \timesten^{-2}$ & $6.28 (\pm 1.39) \timesten^{-1}$ & & $2.15 (\pm 0.39) \timesten^{-2}$ \\
		\bottomrule
		\multirow{3}{*}{Fluorescing DNA} & $V= 0$ (Constant) & $7.93(\pm 0.01) \timesten^{-2}$ & $1.49( \pm 1.89)\timesten^{-2}$ & \multirow{3}{*}{N/A} & $1.05(\pm 0.01)\timesten^{0\phantom{-}}$ \\
			& $V=1$ (Constant) & $1.87(\pm 0.05)\timesten^{-1}$& $8.04(\pm4.03)\timesten^{-2}$ & & $1.00(\pm 0.00) \timesten^{0\phantom{-}}$ \\
			& $V=2$ (Constant) & $2.03(\pm 0.15)\timesten^{-1}$& $2.39(\pm2.09)\timesten^{-1}$ & & $9.45(\pm 0.02) \timesten^{-1}$ \\
		\bottomrule
	\end{tabular}
	\caption{Comparison of the MSE defined in \eqref{eq:orthog_1}, \eqref{eq:orthog_2}, for the proposed method across different datasets. The estimation error for the drift coefficient $\mathcal{L}_\mu$ is reported for learning $\mu$ with an MLP, and compared with the information theoretical CRLB.}
	\label{tab:sde_results}
\end{table*}}

\subsection{Experiment Setup}
\begin{figure}[ht!]
    \centering
    \includegraphics[width=\textwidth, trim={0cm, 0.4cm, 0cm, 0cm},clip]{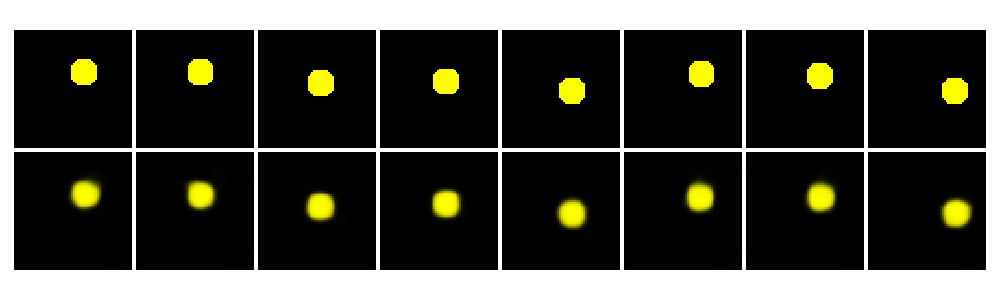}\vspace{-5pt}
    \includegraphics[width=\textwidth, trim={0cm, 0.4cm, 0cm, 0.5cm},clip]{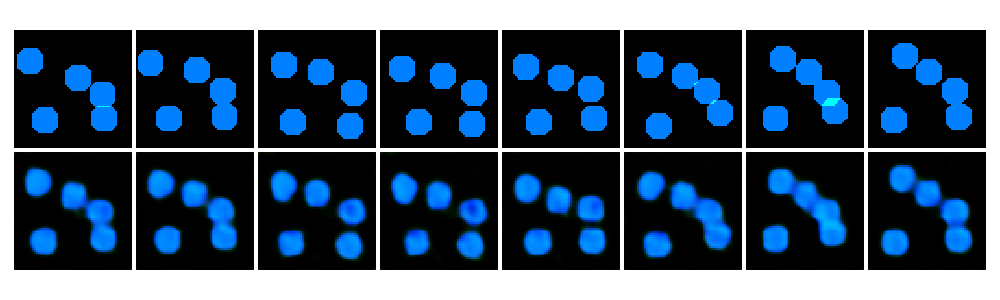}\vspace{-5pt}
    \includegraphics[width=\textwidth, trim={0cm, 0.4cm, 0cm, 0.5cm},clip]{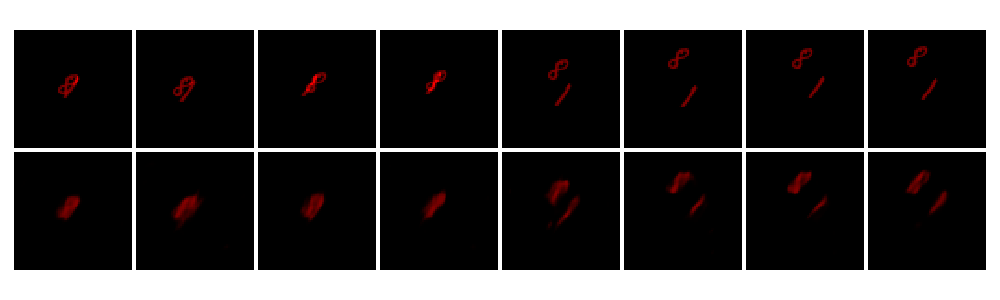}\vspace{-5pt}
    \includegraphics[width=\textwidth, trim={0cm, 0.4cm, 0cm, 0.5cm},clip]{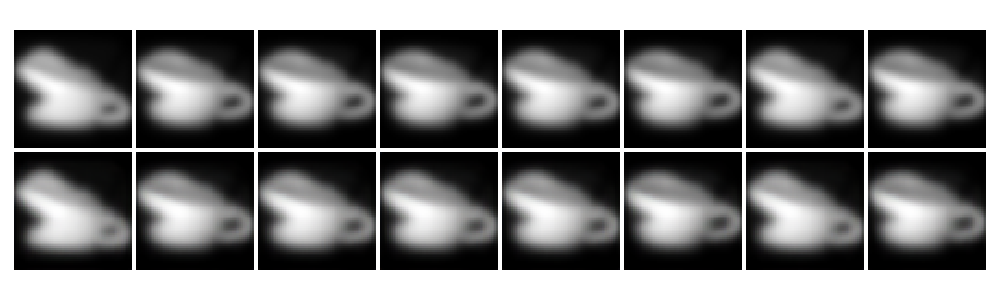}
    \vspace{-5pt}
    \includegraphics[width=\textwidth, trim={0cm, 0.4cm, 0cm, 0.5cm},clip]{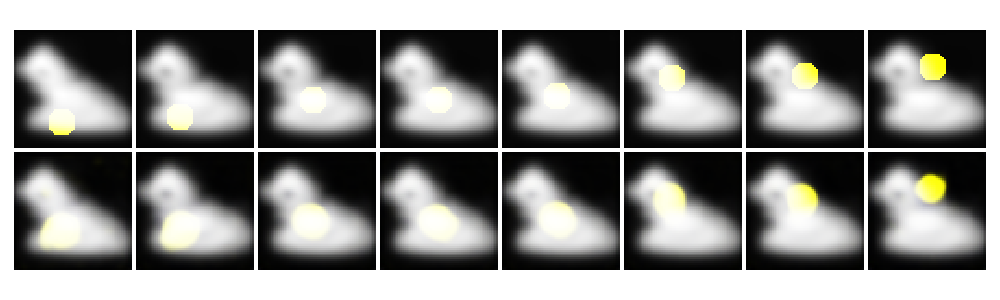}
    \caption{Qualitative comparison of image reconstructions for the Ornstein-Uhlenbeck process on held-out test data. Top row is the original data and bottom row is the reconstruction.}
    \label{fig:reconstruction}
\end{figure}
For all experiments, we use a convolutional encoder-decoder architecture. The latent drift coefficient $\mu$ is represented as a multi layer perceptron (MLP). We consider the same network architectures between all experiments in order to maintain consistency.
All architecture and hyper-parameter specifications are available in the supplemental material.
All image sizes are 64 $\times$ 64 $\times$ 3 making the ambient dimension of size 12,288.

First, we test the proposed method on learning the latent mapping in these datasets. In order to show evidence of the theoretical results presented in Section~\ref{sec:identifiability}, we measure the mean square error (MSE) between the true latent representation and the true drift coefficient, with the ones estimated by the VAE. 
Since Theorems \ref{thm:identifiability_main} and \ref{thm:VAE_convergence} imply that the true latent space and the one obtained by the VAE are equal up to an isometry, we measure the MSE using the following formulas.
\begin{align}
    \label{eq:orthog_1}
	\mathcal{L}_\text{latent} &= \frac{1}{N}\min_{Q,b}\sum_{t = 0 }^N \| Q\tilde f^{-1}(X_t) + b - Z_t \|_2^2, \\
	\mathcal{L}_\mu &= \frac{1}{|\mathcal{X}|}\sum_{x \in \mathcal{X} } \| Q\hat{\mu}(Q^T(x - b)) - \mu(x) \|_2^2.
	\label{eq:orthog_2}
\end{align}
Here, the minimum is over all orthogonal matrices $Q\in \R^{d\times d}$ and vectors $b\in \R^d$, and $\tilde f^{-1}$ is the function learned by the VAE encoder. The minimizers of \eqref{eq:orthog_1} are calculated using a closed form solution which we describe in the Appendix~\ref{sec:crlb}. To calculate \eqref{eq:orthog_2}, we use the minimizers of \eqref{eq:orthog_1}, and $\mathcal{X}$ is the set of sampled points in the run. 
All experiments are repeated over 5 independent runs, and the average and standard deviation are reported in Table~\ref{tab:sde_results}. We also compare $\mathcal{L}_\mu$ obtained by learning $\mu$ using an MLP and a Cram\'er Rao lower bound (CRLB) for $\mathcal{L}_\mu$. The CRLB is obtained using an information theoretic argument, and provides a lower bound for the MSE of any estimator of $\mu$, therefore can considered as a baseline of what is theoretically achievable. The derivation of the CRLB for these experiments is described in Appendix~\ref{sec:crlb}. 

Finally, regarding the oracle in Fig.~\ref{fig:example_mu_learned}, we use the same network architecture for the drift coefficient $\mu$, which is trained by maximizing the log-likelihood of the Euler-Maruyama approximation of the latent SDE.

\subsection{Interpretation of results}
Comparing $\mathcal{L}_\text{latent}$ in Table~\ref{tab:sde_results} between the different experiments, we see that the proposed method is able to learn the latent representation better for the yellow ball, which was expected since this was the simplest dataset.
Comparing $\mathcal{L}_\mu$ with the theoretical CRLB, we observe that our method is able to recover the drift coefficient within the same order of magnitude for the OU process in all datasets, the constant drift process (Brownian random walk) for the digits dataset, and the geometric Brownian motion case for the Wasserstein distance. On the other hand, we believe the performance for the yellow ball dataset, constant process, was hindered by the fact that any solution to that SDE is unbounded, and when we rescale the SDE, so that the coordinates are between $0$ and $1$, we lose information.
Additionally, in the circle cases, the data is largely concentrated in the circular region, but certain jumps from the noise cause the extreme points to be poorly learned, leading to a higher MSE. 
This behavior also is exhibited in the double well case where the bulk of the data is within the wells but regions outside the potential have higher MSEs.
\begin{figure}[ht!]
	\centering
	\includegraphics[width=\textwidth]{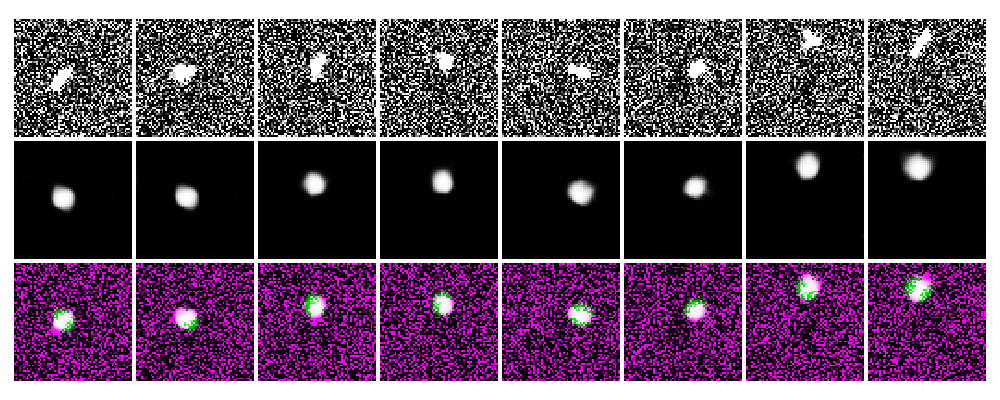}
	\caption{Reconstruction of the DNA molecule on test data from a 2D latent space. The proposed method effectively tracks the location of the molecule while disregarding noise. Top row: histogram normalized data, middle row: reconstruction based on 2D latent space, bottom row: overlay of original input and reconstructed output.}
	\label{fig:dna_sample}
\end{figure}

The final column of Table~\ref{tab:sde_results} describes the MSE between the ground truth image and the reconstruction from the decoder $\|X_t - \hat{f}(\hat{Z}_t) \|_2^2$. 
Examples of the reconstructions are given in Fig.~\ref{fig:reconstruction}, the reconstruction of the test set data qualitatively matches with respect to the locations of the original images.
This suggests that the generative capabilities of the method are effective in generating new images conditioned on proper coordinates given by the latent SDE.

For the noisy synthetic datasets, we compare the MSE to the original, denoised image. 
In these cases, the MSE for the noisy experiments and the noiseless experiments are within one order of magnitude, suggesting the proposed method is effectively denoising the image and tracking only the relevant object governed by the SDE. 
Examples of the denoising are given in Fig.~\ref{fig:noise_recon}. 
For the DNA datasets, we do not have a ground truth and report the MSE between the noisy original images. 
As expected, the MSE is high due to the autoencoder denoising the image. 


\subsection{Adding observation noise}
\begin{figure}
    \centering
    \includegraphics[width=\textwidth]{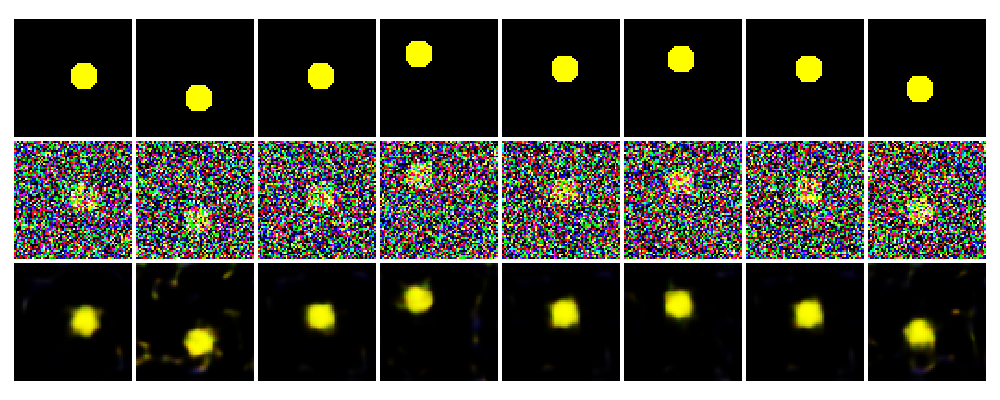}
    \includegraphics[width=\textwidth]{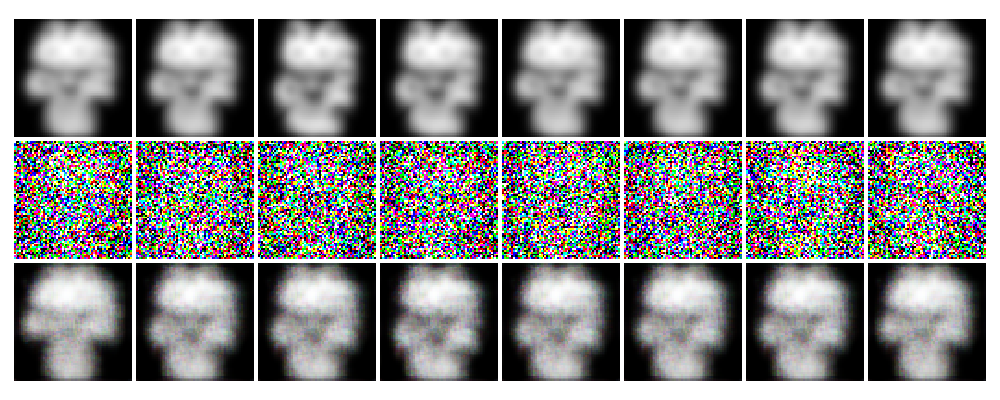}
    \caption{Two experiments with different noise corruption. Top row: original data; middle row: observation; bottom row: reconstruction. Top experiment with $t$-distributed noise, bottom with Gaussian distributed noise.}
    \label{fig:noise_recon}
\end{figure}

In many real applications, such as in the DNA molecule datasets, the observation is corrupted by noise. 
Techniques for dealing with this type of data have been extensively studied in the field of filtering \cite{davis1981introduction}.
We conduct additional experiments where we use the proposed method to uncover latent SDEs with observation noise. 
Although this violates our previous assumption that the observation noise is small, we wanted nevertheless to analyze the empirical performance of our method for more noisy datasets.
For these experiments, we generate the movies according to the same procedures described in previous sections, but we add additional noise to the final output. 
That is, we observe $X_t + \epsilon_t$ where $\epsilon_t$ is sampled either from a Gaussian distribution or a Student's $t$-distribution where the degrees of freedom parameter is set to 3. We include the $t$-distributed noise experiments since the Student's $t$-distribution exhibits a fatter tail, which should make estimation harder.  
We illustrate examples of the image reconstruction capabilities in Fig.~\ref{fig:noise_recon} for the moving ball and Wasserstein datasets with both Student-$t$ and Gaussian noise.

The results suggest that, even though the encoder approximation is valid when the observation noise is low, as stated in Proposition~\ref{proposition:encoder_decomp}, empirically our method performs well for datasets with considerable observation noise: all experiments are within 25\% error of the noise-free experiments.

\subsection{Learning the latent dimension}
\begin{figure}
    \centering
    \includegraphics[width=0.31\textwidth]{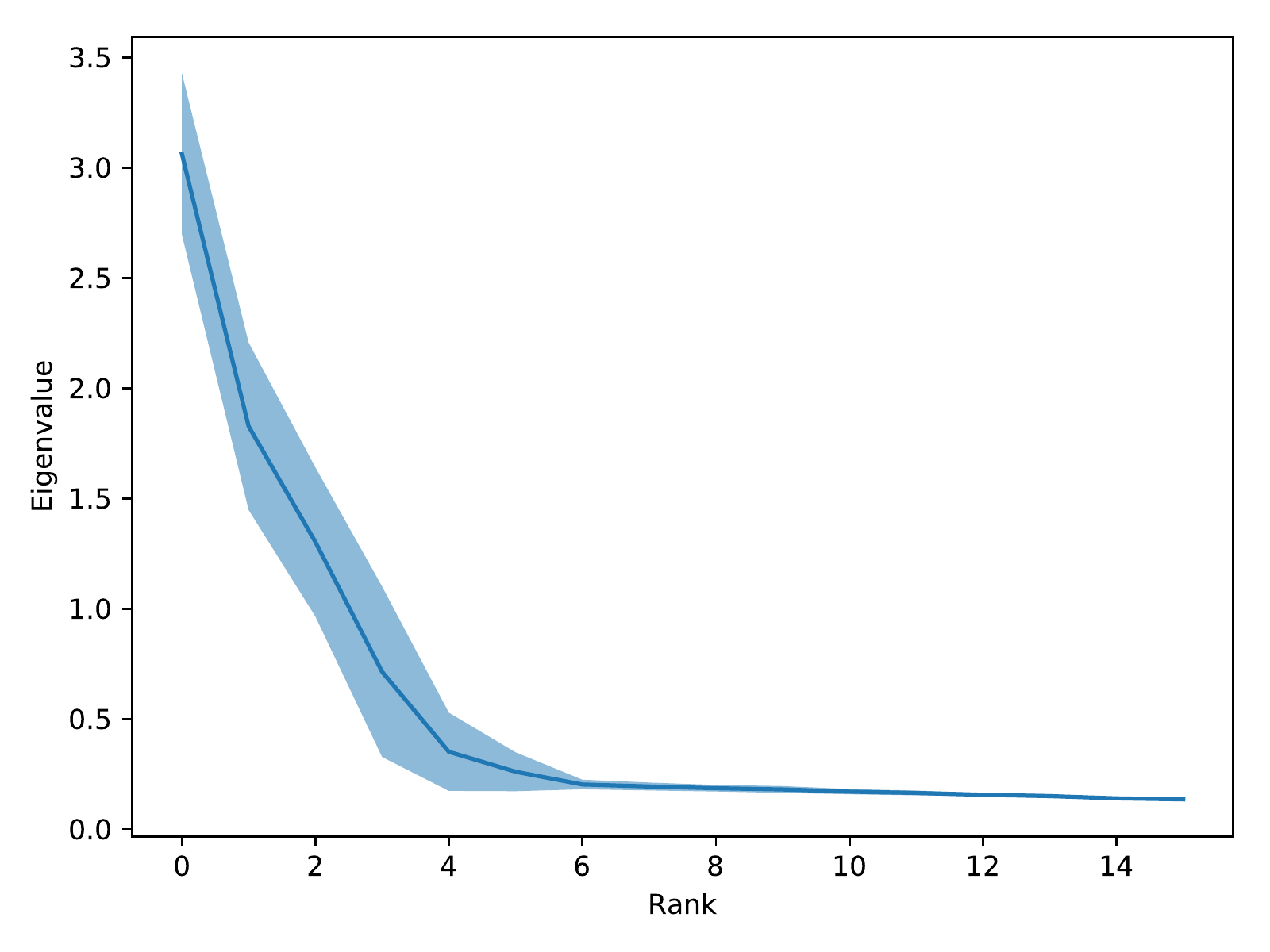}
    \includegraphics[width=0.31\textwidth]{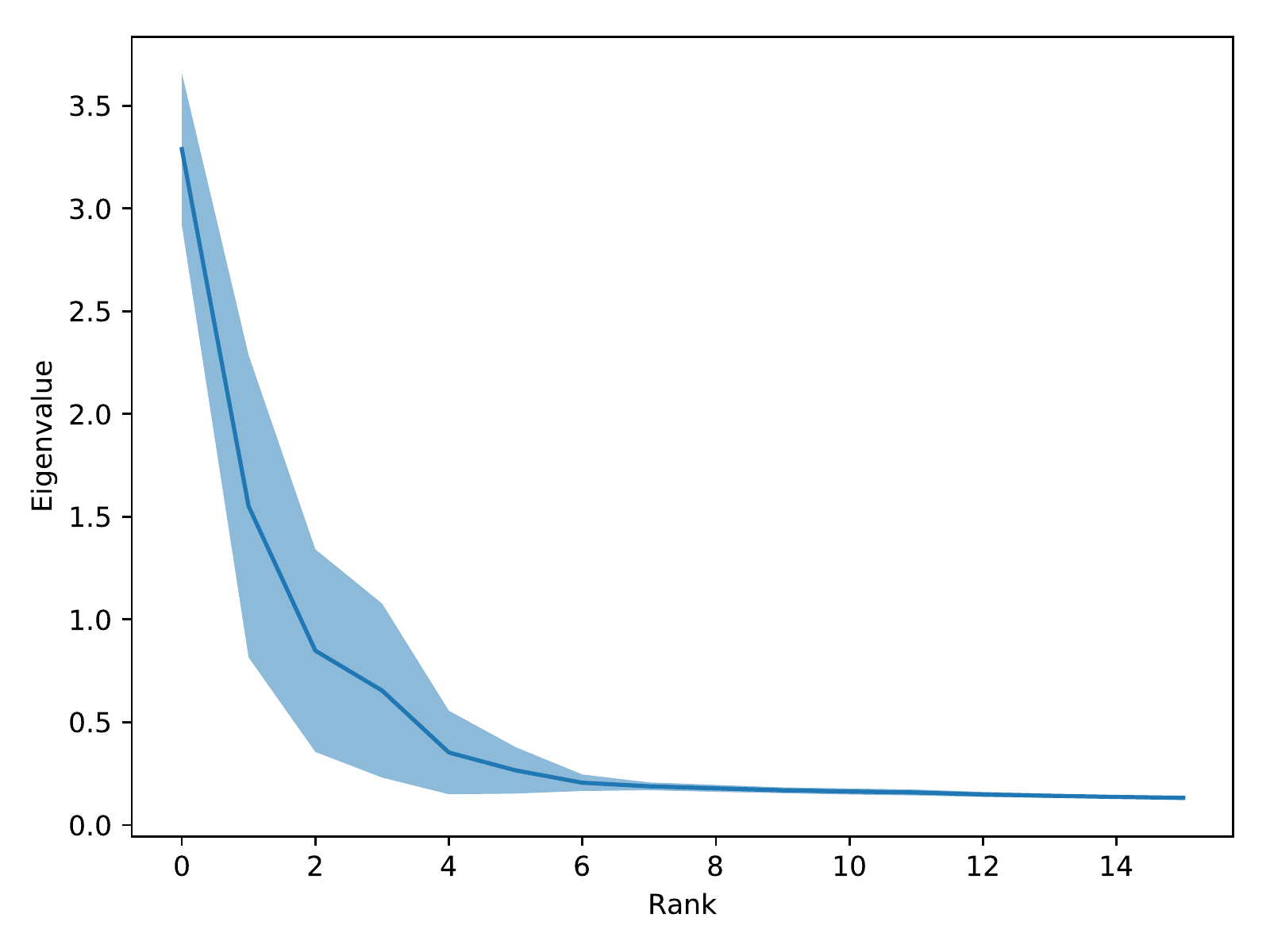}
    \includegraphics[width=0.31\textwidth]{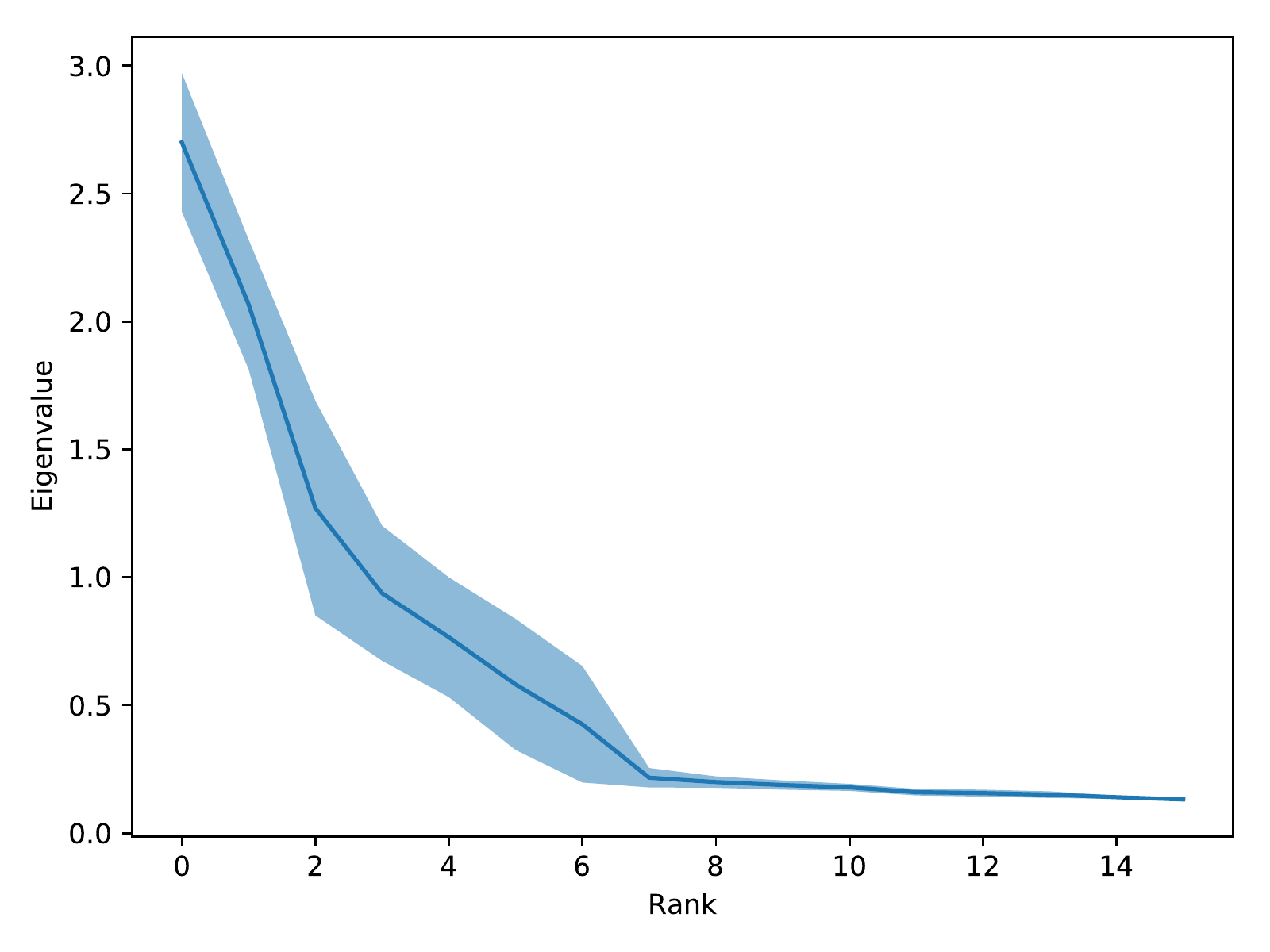}

    \caption{Sorted diagonal entries learned by the heuristic described in Section~\ref{sec:latent_dimension} when applied to the fluorescent DNA datasets \cite{lameh2020controlled} with $V=0$ (left), $V=1$ (middle) and $V=2$ (right). The latent space has dimension 16. }
    \label{fig:latent_size}
\end{figure}

In order to validate the heuristic described in Section~\ref{sec:latent_dimension}, we analyze three additional experiments where we attempt to learn the dimension of the latent space using that heuristic. 
We assume the ground truth is 2-dimensional in all cases, corresponding to the planar movement of the molecule. 
However, there exists additional movement in the orientation of the molecule, which may contribute to the latent dynamics and a latent space larger than 2.
We illustrate the diagonal entries obtained by the heuristic, sorted in decreasing order, in Fig.~\ref{fig:latent_size}.
We use the same architecture and parameters as in the experiments described in Section~\ref{sec:dna}, but for these experiments we include a learnable parameter for the diffusion term and set the latent space dimension to 16. 

The estimated diagonal entries show a rapid decay towards zero, with only the first few having values greater than $0.1$. 
Practically speaking, one would need to choose a threshold for which the diagonal values below this threshold are considered noise and not part of the true latent space. 
\begin{figure}
    \centering
    \includegraphics[width=\textwidth, trim={0cm, 1cm, 0cm, 0cm}, clip]{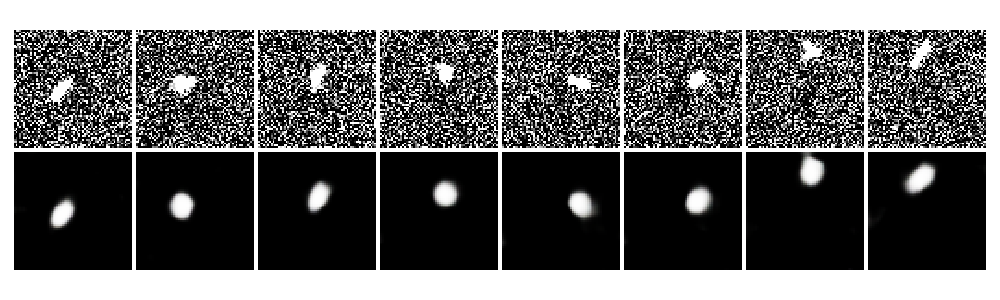}
    \includegraphics[width=\textwidth, trim ={0cm, 1cm, 0cm, 3.8cm}, clip]{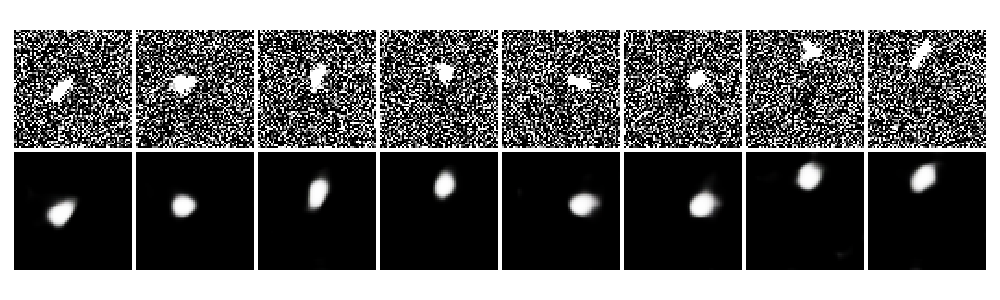}
    \includegraphics[width=\textwidth, trim ={0cm, 0cm, 0cm, 3.8cm}, clip]{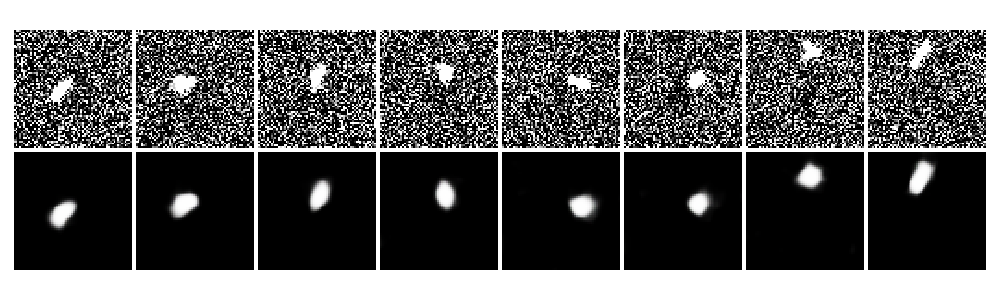}
    \caption{Comparison of reconstructions on test set for different size latent space. Increasing latent size improves reconstruction of orientation. From top to bottom, first row: Original data; second row: 4-D latent space; third row: 6-D latent space; fourth row: 8-D latent space.}
    \label{fig:dna_latent_comp}
\end{figure}

Additional experiments are included in Appendices \ref{sec:comparison_duncker2019learning} and Appendix~\ref{sec:comparison_image_registration}. 

\section{Discussion}\label{sec:discussion}

In this paper we present a novel approach to learn latent SDEs using a VAE framework.
We describe a method that applies to very high dimensional cases, including video data.
We showed that a large class of latent SDEs can be reduced to latent SDEs with isotropic diffusion coefficients. We prove that the proposed method is able to recover a the latent SDE in this class, up to isometry, and validate our results with numerical experiments.
In most cases, the experiments suggest the method is learning the appropriate SDE up to the order of the Cram\'er-Rao lower bounds we obtained, with a few cases being more difficult than others. 

We anticipate the proposed theory and numerical results to lay the foundations for a multitude of downstream applications.
As an example, the proposed method could be used to learn an SDE governing a time series of patient imaging data. The latent drift function could be used to determine whether or not a patient is at high risk for significant deterioration and can help with intervention planning.

There are a variety of additional avenues for expanding the method.  
Recently, the work by \cite{sorrenson2020disentanglement} describes an extension to the framework established by \cite{khemakhem2019variational} wherein the authors propose a method for identifiability that does not require knowledge of the intrinsic dimension. 
A similar approach can be employed in this method where eigenvalues of the estimated diffusion coefficient diminish to form a low rank matrix, indicating unnecessary components. Another extension would be to extend our results to diffusion coefficients that do not satisfy \eqref{eq:no_curl_cond_}.
We can also consider alternative metrics rather than the KL divergence for regularizing the increments, such as the Fisher divergence or Wasserstein distance. 

In other directions, we consider extending our analysis for problems where the samples are sparsely sampled in time.
Specifically, how the changes in an underlying integrator affect the theoretical analysis and how existing integrators such as the one proposed in \cite{roberts2001inference} can be merged into the proposed method.
Furthermore, a thorough analysis of using a neural network to model the drift coefficient, versus parametric forms, or a dictionary of candidate functions, similar to \cite{rudy2017data}, is warranted, in the interest of interpretability.
Moreover, the recovery of the latent SDE provides a natural extension for stochastic control and reinforcement learning where guarantees can be achieved based on the recovered drift function.

A very promising extension of the work is to the case where there is significant observation noise or the observation satisfies an SDE driven by non-Gaussian noise, as is common in filtering problems. A straightforward way to extend our method for this purpose would be a practical implementation of an encoder that decomposes as in \eqref{eq:encoder_decomposition}.
This type of problem was addressed by \cite{duncker2019learning} where they consider observations with point processes and with additional noise corruption. 
Considering such observations can increase the applicability of the proposed method to additional problems where this type of observation noise is prevalent. 

Finally, considering other types of stochastic processes, such as L\'evy flights or jump processes, could provide additional meaningful avenues for applications of the work. 
\appendices

\numberwithin{equation}{section}
\renewcommand{\theequation}{\Alph{section}.\arabic{equation}}
\renewcommand{\thefigure}{\Alph{section}.\arabic{figure}}
\renewcommand{\thetable}{\Alph{section}.\arabic{table}}
\renewcommand{\thetheorem}{\Alph{section}.\arabic{theorem}}
\newcommand{\hA}{{\hat{A}}}
\newcommand{\hX}{{\hat{X}}}
\newcommand{\hmu}{{\hat{\mu}}}
\newcommand{\hS}{{\hat{\Sigma}}}

\renewcommand{\thesection}{\Alph{section}.\hskip-0.6em}
\renewcommand{\thesubsection}{\Alph{section}.\arabic{subsection}\hskip-0.6em}

\section{Calculating the VAE loss}\sectionlabel{sec:VAE_loss}
To calculate the expectation in \eqref{eq:L_def} we use three techniques: Exact formula (when possible), the reparametrization trick \cite{kingma2013auto} and a first order Taylor approximation, which we explain as follows. 
Let $Z$ be a random variable with expectation $\mu$, and suppose we want to estimate $\E[f(Z)]$ for some function $f$. We can use the reparametrization trick to estimate this, but an even simpler way is the rough estimate $\E[f(Z)]\approx f(\mu)$. This corresponds to the exact value of a first order Taylor approximation of $f$:
\begin{align*}
	\E[f(Z)]
	&=\E [f(\mu) + J_f(\mu) (Z - \mu)\\ &\hspace{20pt}+ \{\text{2\textsuperscript{nd} order and higher terms}\} ],\\
	&=f(\mu) + \{\text{2\textsuperscript{nd} order and higher terms}\}.
\end{align*}
Going back to calculating the training loss, we  expand \eqref{eq:L_def}, getting
\begin{align}
	\nonumber \mathcal{L}(\phi,\psi) = \E_{q_{\mathcal{D}}(\bX)}\big[\E_{\tilde q_\psi(\bZ| \bX)}\big[ \log \tilde q_{ \psi}(Z_{t+\deltat}|X_{t+\deltat}) \\   + \log \tilde q_\psi(Z_{t}|X_{t})\label{eq:VAE_loss1} \\
	\label{eq:VAE_loss_prior}
- \log p_{\gamma}(Z_t)\\
- \log p_{\mu}(Z_{t+\deltat}|Z_t)
	\label{eq:VAE_loss2}\\
	  - \log p_f(X_{t+\deltat}|Z_{t+\deltat}) - \log p_f(X_t| Z_t)
	\big]\big] \label{eq:VAE_loss3}
\end{align}
We now explain how we calculate/estimate each of the expectations \eqref{eq:VAE_loss1}, \eqref{eq:VAE_loss_prior}, \eqref{eq:VAE_loss2}, and \eqref{eq:VAE_loss3}. As in vanilla VAEs, our encoder $\tilde q_\psi$ is given by a Gaussian distribution conditioned on $X_t$.
We define $\bm{\mu}_{\tilde q}(X_t)$ and $\bm{L}_{\tilde q}(X_t)$ as the neural networks that encode its mean and its Cholesky decomposition of the covariance matrix. That is, letting $\bm{\Sigma}_{\tilde q}(X_t)$ be the covariance matrix, we have $\bm{\Sigma}_{\tilde q}(X_t) = \bm{L}_{\tilde q}(X_t) \bm{L}_{\tilde q}(X_t)^\top$. The probability distribution function $\tilde q_\psi$ is defined by:
\begin{align*}
	&\tilde q_\psi(Z_{t}|X_{t}) =\\& \frac1{\sqrt{2\pi}^d |\det \bm{L}_{\tilde q}(X_t)|} \exp\left(-\frac{\|\bm{L}_{\tilde q}(X_t)^{-1} (Z_t - \bm{\mu}_{\tilde q}(X_t))\|^2}2 \right).
\end{align*}
This implies $\E[Z_{t}|X_t] = \bm{\mu}_{\tilde q}(X_t)$ and 
\begin{align*}
\E[Z_{t}Z_{t}^\top|X_t] &= \bm{L}_{\tilde q}(X_t)\bm{L}_{\tilde q}(X_t)^\top + \bm{\mu}_{\tilde q}(X_t)\bm{\mu}_{\tilde q}(X_t)^\top,\\
&= \bm{\Sigma}_{\tilde q}(X_t) + \bm{\mu}_{\tilde q}(X_t)\bm{\mu}_{\tilde q}(X_t)^\top,
\end{align*}thus \eqref{eq:VAE_loss1} can be calculated exactly\footnote{We do not include the term $\sqrt{2\pi}^d$ since it is a constant that does not influence the optimization.}:
\begin{align*}
	&\mathscalebox{.89}{\E_{q_\psi(\bZ| \bX)}\left[ \log q_\psi(Z_{t+\deltat}|X_{t+\deltat})|\bX\right]}\\
    &=\mathscalebox{.89}{-\log |\det \bm{L}_{\tilde q}(X_t)|}\\ &
	\mathscalebox{.89}{-\frac12 \E\left[\tr\left((Z_t\hspace{-1pt} - \hspace{-1pt}\bm{\mu}_{\tilde q}(X_t))
	(Z_t\hspace{-1pt} - \hspace{-1pt} \bm{\mu}_{\tilde q}(X_t))^\top \bm{L}_{\tilde q}(X_t)^{-\top} \bm{L}_{\tilde q}(X_t)^{-1} \right)|X_t\right]\hspace{-1pt},}\\
	&=\mathscalebox{.89}{-\log |\det \bm{L}_{\tilde q}(X_t)|-\frac12 \tr\left(\bm{\Sigma}_{\tilde q}(X_t) \bm{\Sigma}_{\tilde q}(X_t)^{-1} \right),}\\
	&=\mathscalebox{.89}{-\log |\det \bm{L}_{\tilde q}(X_t)| -\frac{d}2 .}
\end{align*}
Regarding \eqref{eq:VAE_loss_prior}, we observed that having the prior distribution $p_\gamma$ to be Gaussian, with a fixed isotropic covariance controlled by an hyper-parameter $\nu$, worked best empirically. In this case, the expectation can then be calculated exactly:
\begin{align*}
\hspace{50pt}&\hspace{-50pt}\E[-\log p_{\gamma}(Z_t)|X_t] 
\\&= -\frac{d}2\log \nu + \frac{\nu}2 \E\left[\|Z_t\|^2|X_t\right],
\\&= -\frac{d}2\log \nu + \frac{\nu}2  \tr\left(\bm{\Sigma}_{\tilde q}(X_t)\right) + \frac{\nu}2 \|\bm{\mu}_{\tilde q}(X_t)\|^2.
\end{align*}
For calculating the loss we can ignore the term $\frac{d}2\log \nu$, since it stays constant during training.
The term \eqref{eq:VAE_loss2} is equal to:
\begin{align*}
	&\E_{q_\psi(\bZ| \bX)}\left[ -\log p_{\mu}(Z_{t+\deltat}|Z_t)| \bX\right]\\
	&=\frac{d}2\log \deltat +\frac1{2\Delta t} \E\Big[\tr\big(\left(Z_{t+\deltat} - Z_t - \mu(Z_t) \deltat\right) \\& \hspace{100pt}\left(Z_{t+\deltat} - Z_t - \mu(Z_t) \deltat\right)^\top\big)| \bX\Big].
\end{align*}
We note that the only expectations that cannot be calculated exactly are the ones involving $\mu(Z_t)$. Conditioned on $X_t$, the distribution of $Z_t$ is given by the encoder $q_\psi$, thus we use the first order approximation $\mu(Z_t)\approx \mu(\bm{\mu}_{\tilde q}(X_t))$. 
Calculating the other expectations, and noting that the definition $q_\psi(\bZ| \bX)$ implies that $\Cov(Z_t,Z_{t+\Delta t})=0$, we get the formula:
\begin{align*}
	&\E_{q_\psi(\bZ| \bX)}\left[ -\log p_{\mu}(Z_{t+\deltat}|Z_t)\right]\\
	&=\frac{d}2\log \deltat +\frac1{2\Delta t}\tr(\bm{L}_{\tilde q}(X_t)\bm{L}_{\tilde q}(X_t)^\top) \\
	&\hspace{50pt}+ \frac1{2\Delta t}\tr(\bm{L}_{\tilde q}(X_{t+\Delta t})\bm{L}_{\tilde q}(X_{t+\Delta t})^\top) \\
	&\hspace{50pt}+ \frac1{2\Delta t}\|\bm{\mu}_{\tilde q}(X_{t+\deltat}) - \bm{\mu}_{\tilde q}(X_{t}) - \mu(\bm{\mu}_{\tilde q}(X_{t}))\|^2.
\end{align*}
Finally we model the noise $\epsilon$ in \eqref{eq:VAE_loss3} as a centered Gaussian random variable with variance $\tau I$, where $\tau$ is a hyper-parameter. We have,
\begin{equation*}
\mathscalebox{.94}{\E_{q_\psi(\bZ| \bX)}\left[-\log p_f(X_t| Z_t)\right]=\frac{d}2\log \tau +\frac1{2\tau}\E\left[\|f(Z_t)- X_t\|^2\right],}
\end{equation*}
and use the reparametrization trick to calculate this. 
\section{Proof of Lemmas}
\subsection{Proof of Lemma \ref{lemma:whatcansigmabe}}\label{sec:sigma_lemma_proof}
\begin{proof}
The implications $\textit{(\ref{lemma:whatcansigmabe}.i)}\Rightarrow \textit{(\ref{lemma:whatcansigmabe}.ii)} \Rightarrow \textit{(\ref{lemma:whatcansigmabe}.iii)}$ are trivial and the Brownian motion is a particular case of \textit{(\ref{lemma:whatcansigmabe}.iii)}, thus we show that $\textit{(\ref{lemma:whatcansigmabe}.iii)} \Rightarrow \textit{(\ref{lemma:whatcansigmabe}.iv)}$ and $\textit{(\ref{lemma:whatcansigmabe}.iv)}\Leftrightarrow\textit{(\ref{thm:canonical_sigma_}.ii)}$. 
To show $\textit{(\ref{lemma:whatcansigmabe}.iii)}\Rightarrow \textit{(\ref{lemma:whatcansigmabe}.iv)}$, notice that for all $y\in\R^d$, $\sigma^{-1}(y) = M^{-\top} \varLambda(M^{-1}y)^{-1} M^{-1}$. Since $\varLambda$ satisfies condition \textit{(\ref{lemma:whatcansigmabe}.ii)}, we have $\varLambda(y)$ is a diagonal matrix and letting $\tilde m_i$ be the $i$-th row of $M^{-1}$, $\varLambda(M^{-1} y)_{ii} = f_{i}(\tilde m_i^\top y)$ for all $y\in \R^d$ and $i\in \{1,\dots,d\}$. Since $\sigma^{-1}(y) = M^{-\top} \varLambda(M^{-1}y)^{-1} M^{-1}$, and  and $\varLambda$ is diagonal, we have,
\begin{align*}
\sigma^{-1}(y) &= \sum_{i=1}^d \varLambda(M^{-1} y)_{ii} \tilde m_i \tilde m_i^\top,\quad \forall y\in \R^d  \\
&= \sum_{i=1}^d f_{i}(\tilde m_i^\top y) \tilde m_i \tilde m_i^\top,\quad \forall y\in \R^d    
\end{align*}
Since $f_{i}$ is positive, there is a convex function $g_i$ such that $g_i'' = f_i$, which implies that $f_{i}(\tilde m_i^\top y) \tilde m_i \tilde m_i^\top$ is the Hessian of $g_i(\tilde m_i^\top y)$. $\sigma^{-1}$ is then the hessian of a positive linear combination of convex functions, which is thus a convex function.
To show that $\textit{(\ref{thm:canonical_sigma_}.ii)}\Rightarrow \textit{(\ref{lemma:whatcansigmabe}.iv)}$, we note that \eqref{eq:no_curl_cond} implies that for all $j,k$
	\begin{align*}
		\frac{\partial}{\partial y_k} \left(\sigma(y)^{-1}\right) e_j 
		&= - \sigma(y)^{-1} \frac{\partial\sigma(y)}{\partial y_k} \sigma(y)^{-1} e_j  \\
		&= - \sigma(y)^{-1} \frac{\partial\sigma(y)}{\partial y_j} \sigma(y)^{-1} e_k, \\
		&= \frac{\partial}{\partial y_j} \left(\sigma(y)^{-1}\right)  e_k,
	\end{align*}
	thus \cite[Theorem 11.49]{lee2013smooth} implies that there exists a function $h:\R^d\to \R^d$ such that $\sigma(y)^{-1} = J_h(y)$ for all $y\in \R^d$, where $J_h$ is the Jacobian of $h$. Moreover, since $\sigma^{-1}(y)$ is symmetric,
	\begin{align*}
		\frac{\partial}{\partial y_j} h_i(y)
		&= (\sigma(y)^{-1})_{ij}= (\sigma(y)^{-1})_{ji}= \frac{\partial}{\partial y_i} h_j(y),
	\end{align*}
	thus applying \cite[Theorem 11.49]{lee2013smooth} again, there is a function $g:\R^d\to \R$ such that $h(y) = \nabla g(y)$, which implies then that $\sigma(y)^{-1} = H_g(y)$, and since $\sigma(y)^{-1}$ is positive definite, $g$ is convex. To show the other direction, suppose that $\sigma(y)^{-1} = H_g(y)$, then, for all $i,j,k\in \{1,\dots,d\}$,
	\begin{align*}
		\frac{\partial}{\partial y_k} \left(\sigma(y)^{-1}\right)_{ij}
		&= \frac{\partial^3 g(y)}{\partial y_i \partial y_j \partial y_k },\\
		&= \frac{\partial}{\partial y_j} \left(\sigma(y)^{-1}\right)_{ik},
	\end{align*}
	which finally implies 
	\begin{align*}
		\frac{\partial\sigma(y)}{\partial y_k} \sigma(y)^{-1} e_j
		&= - \sigma(y)\frac{\partial}{\partial y_k} \left(\sigma(y)^{-1}\right) e_j\\
        &= - \sigma(y)\frac{\partial}{\partial y_j} \left(\sigma(y)^{-1}\right) e_k\\
        &=\frac{\partial\sigma(y)}{\partial y_j} \sigma(y)^{-1} e_k
	\end{align*}
\end{proof}
\vspace{-28pt}
\subsection{Proof of Proposition~\ref{proposition:encoder_decomp}}\label{sec:encoder_proof}

First, we recall the definition of mutual information.
The mutual information between the random variables $X$ and $Y$, with joint distribution $p_{X,Y}$, marginal distributions $p_X$ and $p_Y$, respectively, and conditional distribution $p_{Y|X}$, is defined as:
\begin{align}
	\nonumber I(X;Y)&:= D_{KL}(p_{X,Y}(X,Y)\;\|\;p_{X}(X)p_Y(Y)),\\
	&= \E_{p_{X}(X)}\left[D_{KL}(p_{Y|X}(Y|X)\;\|\;p_Y(Y))\right].
\end{align}
We now proceed with the proof of Proposition~\ref{proposition:encoder_decomp}.
By the chain rule of mutual information, the following identities hold:
\begin{align}
	\nonumber \hspace{20pt}&\hspace{-20pt} I(Z_{t+\deltat};Z_t|X_{t+\deltat}) \\ \nonumber & =  I(Z_{t+\deltat},X_{t+\deltat};Z_t) - I(X_{t+\deltat};Z_t),\\
	\nonumber &\le I(Z_{t+\deltat},X_{t+\deltat};Z_t),\\
	&= I(Z_{t+\deltat};Z_t) +I(X_{t+\deltat};Z_t|Z_{t+\deltat}).\label{eq:VAE_MI_eq}
\end{align}
Note that \eqref{eq:Xt_factorization} implies that $X_{t+\deltat}$ and $Z_t$ are conditionally independent given $Z_{t+\deltat}$, which implies that $I(X_{t+\deltat};Z_t|Z_{t+\deltat})=0$ and 
$I(Z_{t+\deltat};Z_t|X_{t+\deltat}) \le I(Z_{t+\deltat};Z_t) \ll I(Z_{t+\deltat};X_{t+\deltat}).$
Finally, applying the chain rule,
\begin{align*}
	\nonumber I(Z_{t+\deltat};X_{t+\deltat})  \hspace{-3pt} &\hspace{2pt}\le \hspace{-1pt} I(Z_{t+\deltat},X_{t+\deltat};Z_t), \\
&= I(Z_{t+\deltat};X_{t+\deltat}) +  I(Z_{t+\deltat};Z_t|X_{t+\deltat}), \\
&\le I(Z_{t+\deltat};X_{t+\deltat}) + I(Z_{t+\deltat};Z_t),\\
&\approx I(Z_{t+\deltat};X_{t+\deltat}).
\end{align*}
\vspace{-10pt}
\section{Proof of Theorem \ref{thm:identifiability_main}}\sectionlabel{sec:proof_thm3}
Our proof is very similar to the proof of \cite[Theorem 1]{khemakhem2019variational}. However, we cannot use that result directly, since our generative model is slightly different. Therefore we replicate the proof here, adapted to our generative model, but suggest consulting \cite{khemakhem2019variational} to understand more intricate details of the proof.

Our proof is split in 3 steps:
\begin{enumerate}[(I)]
\item First, we write our generation model as a convolution that depends on the noise $\epsilon$, to reduce the equality $p_{\theta}(\mathbf x) = p_{\theta^*}(\mathbf x)$ to the noiseless case, obtaining \eqref{eq:noiseless_p_equality}.

\item Next, we use some algebraic tricks to get a linear dependence between $f_*^{-1}$ and $f^{-1}$, obtaining \eqref{eq:ftildefcorrespondence}.

\item Finally, we keep using algebraic manipulations to get the rest of the statements in the theorem.
	
\end{enumerate}

\subsubsection*{Step (I)}

Let $\cZ=\R^d$ be the latent space, $\cX=f(\cZ)$, that is, $x\in \cX$ if there is $z\in\cZ$ such that $f(z)=x$. First, we write $p_{\theta}(x_{t + \Delta t}, x_t)$ as a convolution.
\begin{align}
\nonumber & p_{\theta}(x_{t + \Delta t}, x_t) \hspace{-.1\linewidth}  \\
\tag{i} \label{eq:ident_thm_step1_1} &= \int_{\cZ}\int_{\cZ} p_\epsilon(x_{t + \Delta t} - f(z_{t + \Delta t})) p_\epsilon(x_t - f(z_t)) \\ \nonumber & \hspace{60pt} q_{\mu, \gamma}(z_{t + \Delta t},z_t) \d z_{t + \Delta t} \d z_{t}, \\
\tag{ii} \label{eq:ident_thm_step1_2}&= \int_{\cX}\int_{\cX} p_\epsilon(x_{t + \Delta t} - \tilde x_{t + \Delta t}) p_\epsilon(x_t - \tilde x_t) \\ \nonumber & \hspace{60pt} q_{\mu, \gamma, f^{-1}}(\tilde x_{t + \Delta t}, \tilde x_{t})  \d \tilde x_{t + \Delta t} \d \tilde x_{t}, \\
\tag{iii} \label{eq:ident_thm_step1_3} &= \int_{\R^n}\int_{\R^n} p_\epsilon(x_{t + \Delta t} - \tilde x_{t + \Delta t}) p_\epsilon(x_t - \tilde x_t) \\ \nonumber & \hspace{60pt} q_{\mu, \gamma, f^{-1}, \cX}(\tilde x_{t + \Delta t}, \tilde x_{t})  \d \tilde x_{t + \Delta t} \d \tilde x_{t}, \\
\tag{iv} \label{eq:ident_thm_step1_4} &= \left[(p_\epsilon \times p_\epsilon) \ast  q_{\mu, \gamma, f^{-1}, \cX}\right](\tilde x_{t + \Delta t}, \tilde x_{t}),
\end{align}
wherein:%
\begin{enumerate}
\item[\eqref{eq:ident_thm_step1_1}] We write $p_{\theta}(x_{t+\Delta t}, x_t)$ as the integration over $z_t,z_{t + \Delta t}$ of $p_{\theta}(x_{t + \Delta t}, x_t, z_{t + \Delta t}, z_t)$, expand $p_{\theta}$ as in \eqref{eq:Xt_factorization}, and let 
\begin{equation*}
q_{\mu, \gamma}(z_{t + \Delta t},z_t)=p_{\mu}(z_{t+\deltat}|z_t) p_\gamma(z_t).
\end{equation*}%
\item[\eqref{eq:ident_thm_step1_2}] We do the change of variable $\tilde x_t=f(z_t)$ and similarly for $\tilde x_{t+\deltat}$. As in \cite{khemakhem2019variational}, the change of variable volume term for $\tilde x_t$ is
\begin{equation*}
\vol J_{f^{-1}}(\tilde x_{t}) :=\sqrt{\det \left(J_{f}(f^{-1}(\tilde x_{t}))^\top J_{f}(f^{-1}(\tilde x_{t}))\right)},
\end{equation*}
and $\tilde x_{t+\Delta t}$ is analogous. We then let
\begin{multline*}
\hspace{-25pt} q_{\mu, \gamma, f^{-1}}(\tilde x_{t + \Delta t}, \tilde x_{t})  \\ \hspace{-18pt} = \vol J_{f^{-1}}(\tilde x_{t}) \vol J_{f^{-1}}(\tilde x_{t+\deltat}) q_{\mu, \gamma}(f^{-1}(\tilde x_{t + \Delta t}), f^{-1}(\tilde x_{t})).
\end{multline*}%
\item[\eqref{eq:ident_thm_step1_3}] We let
\begin{multline*}
q_{\mu, \gamma, f^{-1}, \cX}(\tilde x_{t + \Delta t}, \tilde x_{t}) \\ = \mathds{1}_\cX(\tilde x_{t + \Delta t}) \mathds{1}_\cX(\tilde x_{t}) q_{\mu, \gamma, f^{-1}}(\tilde x_{t + \Delta t}, \tilde x_{t}),
\end{multline*}%
so that the domain of the integral may be $\R^n$ instead of $\cX$.
\item[\eqref{eq:ident_thm_step1_4}] We finally notice the convolution formula, with $(p_\epsilon \times p_\epsilon)(\tilde x_{t + \Delta t}, \tilde x_{t})=p_\epsilon(\tilde x_{t + \Delta t})p_\epsilon(\tilde x_{t})$.
\end{enumerate}
We now have that if $p_{\theta}(x_{t + \Delta t}, x_t)=p_{\theta^*}(x_{t + \Delta t}, x_t)$ for all $x_{t + \Delta t}, x_t\in \R^n$, then by taking the Fourier transform we get that for all $x_{t + \Delta t}, x_t\in \R^n$ (or $\omega_{t + \Delta t}, \omega_t\in \R^n$ when applicable)
\begin{align*}
& \left[(p_\epsilon \times p_\epsilon) \ast q_{\mu, \gamma, f^{-1}, \cX}\right](x_{t + \Delta t}, x_{t}) \\ &= \left[(p_\epsilon \times p_\epsilon) \ast q_{\mu_*, \gamma_*, f_*^{-1}, \cX_*}\right](x_{t + \Delta t}, x_{t}),\\
& \mathcal{F}q_{\mu, \gamma, f^{-1}, \cX}(\omega_{t + \Delta t}, \omega_{t})\varphi_\epsilon(\omega_{t + \Delta t}) \varphi _\epsilon(\omega_{t}) \\  &=  \mathcal{F}q_{\mu_*, \gamma_*, f_*^{-1}, \cX_*}(\omega_{t + \Delta t}, \omega_{t})\varphi_\epsilon(\omega_{t + \Delta t}) \varphi_\epsilon(\omega_{t}),\\
& \mathcal{F}q_{\mu, \gamma, f^{-1}, \cX}(\omega_{t + \Delta t}, \omega_{t}) =  \mathcal{F}q_{\mu_*, \gamma_*, f_*^{-1}, \cX_*}(\omega_{t + \Delta t}, \omega_{t}),\\
& q_{\mu, \gamma, f^{-1}, \cX}(x_{t + \Delta t}, x_{t}) =
q_{\mu_*, \gamma_*, f_*^{-1}, \cX_*}(x_{t + \Delta t}, x_{t}),
\end{align*}
where $\mathcal{F}$ denotes Fourier Transform and $\varphi_\epsilon = \mathcal{F}p_\epsilon$ is the characteristic function of $\epsilon$. Here condition 1 guarantees that we can divide by $\varphi_\epsilon(\omega_{t + \Delta t})\varphi _\epsilon(\omega_{t})$. This equality for all $x_t, x_{t + \Delta t}$ implies that $\cX=f(\cZ)=f^*(\cZ)=\cX^*$ and, for all $x_t, x_{t + \Delta t}$ in $\cX$,
\begin{align}
\nonumber &   \log\vol J_{f_*^{-1}}(x_{t + \Delta t}) + \log\vol J_{f_*^{-1}}(x_{t}) \\ \nonumber & + \log p_{\mu_*}(f_*^{-1}(x_{t + \Delta t})|f_*^{-1}(x_{t})) +  \log p_{\gamma_*}(f_*^{-1}(x_{t}))
\\ \nonumber & = \log\vol J_{f^{-1}}(x_{t + \Delta t}) +\log\vol J_{f^{-1}}(x_{t}) \\ & + \log p_\mu(f^{-1}(x_{t + \Delta t})|f^{-1}(x_{t})) +  \log p_\gamma(f^{-1}(x_{t})).
\label{eq:noiseless_p_equality}
\end{align}
\subsubsection*{Step (II)}
We have 
\begin{multline*}
\log p_\mu(f^{-1}(x_{t + \Delta t})|f^{-1}(x_{t})) \\ = -\frac{\|f^{-1}(x_{t + \Delta t}) - f^{-1}(x_{t}) - \mu(f^{-1}(x_{t}))\deltat\|^2}{2 \deltat}. 
\end{multline*}
Let
\begin{equation}\label{eq:lambda_def}
\lambda(x_{t})= f^{-1}(x_{t}) - \mu(f^{-1}(x_{t}))\deltat,
\end{equation}
\begin{equation}\label{eq:psi_def}
\psi(x_{t}) = \vol J_{f^{-1}}(x_{t}) + \log p_\gamma(f^{-1}(x_{t})) - \frac{\|\lambda(x_{t})\|^2}{2 \deltat},
\end{equation}
and
\begin{equation}\label{eq:varphi_def}
\varphi(x_{t + \Delta t}) = \vol J_{f^{-1}}(x_{t + \Delta t}) - \frac{\|f^{-1}(x_{t + \Delta t})\|^2}{2 \deltat},
\end{equation}
and define analogously $\lambda_*, \psi_*$ and $\varphi_*$. We have for all $x_t, x_{t + \Delta t}$ in $\cX$,
\begin{multline}\label{eq:dot_x_t_canonical}
\psi(x_{t}) + \frac1{\deltat}\left< f^{-1}(x_{t + \Delta t}), \lambda(x_{t})\right> + \varphi(x_{t + \Delta t}) \\ = \psi_*(x_{t}) + \frac1{\deltat}\left<f_*^{-1}(x_{t + \Delta t}), \lambda_*(x_{t})\right> + \varphi_*(x_{t + \Delta t}).
\end{multline}
Let $w_0$ be some element in $\cX$, then subtracting two equations we obtain
\begin{multline}\label{eq:dot_x_tw_0}
\left< f^{-1}(x_{t + \Delta t}), \lambda(x_{t}) - \lambda(w_{0})\right> + \zeta(x_{t}) - \zeta(w_{0}) \\ = \left<f_*^{-1}(x_{t + \Delta t}), \lambda_*(x_{t}) -  \lambda_*(w_{0})\right>, 
\end{multline}
where $\zeta(x) = (\psi(x_t) - \psi_*(x_{t}))\deltat$. Let $z_0=f_*^{-1}(w_0)$ and $z_1, \dots, z_d$ a set of vectors in $\cZ$ such that $z_1-z_0, \dots, z_d-z_0$ are linearly independent. Let $\mathbf Z$ the $d\times d$ matrix such that its $i$-th column is $z_i-z_0$ (by the hypothesis, $\mathbf Z$ is an invertible matrix). Moreover let $\mathbf M_*$ the $d\times d$ matrix with $i$-th column defined by $\mathbf{M}_i = \mu_*(z_i) - \mu_*(z_0)$, and finally let $\mathbf{\Lambda}_*$ be the matrix with $i$-th column defined by $\mathbf{\Lambda_*}_i = \lambda_*(f_*(z_{i})) - \lambda_*(w_{0})$. By construction we have $\mathbf{\Lambda}_* = \mathbf{Z} - \deltat \mathbf{M}_*$, or equivalently 
\begin{equation}
\mathbf{\Lambda_* Z}^{-1}= I - \deltat \mathbf{M_* Z}^{-1}.
\end{equation}
Therefore $\mathbf{\Lambda Z}^{-1}$ is singular only if $\frac1{\deltat}$ is an eigenvalue of $\mathbf{M_* Z}^{-1}$. Since that set of eigenvalues is finite, we set $S$ in the theorem statement to be
\begin{equation}
S=\left\{\lambda:\frac1\lambda \text{ is a positive eigenvalue of }\mathbf{M _* Z}^{-1}\right\}
\end{equation}
If now $\deltat\in \R^+\backslash S$, the matrix $\mathbf{\Lambda_* Z}^{-1}$ is non-singular, which implies $\mathbf{\Lambda}_*$ is non-singular. Let $\mathbf{\Lambda}_*$ defined as above, define ${\mathbf{\Lambda}}$ analogously for $\lambda$, and let $\bm{\psi}\in \R^d$ such that 
\begin{equation*}
\bm{\psi}_i = \zeta(f_*(z_{i})) - \zeta(w_0).
\end{equation*}
We write \eqref{eq:dot_x_tw_0} in matrix notation for $x_t=f(z_0),\dots,f(z_d)$.
\begin{equation}
 {\mathbf{\Lambda}}_*^\top f_*^{-1}(x_{t + \Delta t}) = \mathbf{\Lambda}^\top f^{-1}(x_{t + \Delta t}) + \bm{\psi},
\end{equation}
which implies that
\begin{equation}
f_*^{-1}(x_{t + \Delta t}) = \mathbf{\Lambda}_*^{-\top} {\mathbf{\Lambda}}^\top f^{-1}(x_{t + \Delta t}) + \mathbf{\Lambda}_*^{-\top}\bm{\psi},
\end{equation}
holds for all $x_{t + \Delta t}\in \cX$. Equivalently, for all $z\in \cZ$, $f_*(z)\in \cX$ and
\begin{equation}
z = \mathbf{\Lambda}_*^{-\top} {\mathbf{\Lambda}}^\top f^{-1}(f_*(z)) + \mathbf{\Lambda}_*^{-\top}\bm{\psi},
\end{equation}
If we take the derivatives on $z$ on both sides, we get
\begin{equation}
I = \mathbf{\Lambda}_*^{-\top} {\mathbf{\Lambda}}^\top J_{f^{-1} \circ f_*} (z).
\end{equation}
Since the matrix $I$ is non-singular, this implies $\mathbf{\Lambda}_*^{-\top} {\mathbf{\Lambda}}^\top$ is a non-singular matrix. Therefore there exists an invertible matrix $Q$ and a vector $b$ such that
\begin{equation}\label{eq:ftildefcorrespondence}
 f_*^{-1}(x_{t + \Delta t}) = Q f^{-1}(x_{t + \Delta t}) + b.
\end{equation}
\subsubsection*{Step (III)}
Equation \eqref{eq:ftildefcorrespondence} implies that for all $z\in \cZ$
\begin{align*}
f(z)&=f_*(f_*^{-1}(f(z))),\\
&=f_*(Q f^{-1}(f(z)) + b),\\
&=f_*(Q z + b).
\end{align*}
Moreover $J_{f}=J_{f_*} Q$, and $\log\vol J_{f^{-1}} = \log\vol J_{f_*^{-1}} + \log|\det Q|$. Now replacing $x_{t + \Delta t}$ by $f(z)$ on \eqref{eq:dot_x_t_canonical} and using \eqref{eq:ftildefcorrespondence} and \eqref{eq:psi_def}, we get for all $z\in \cZ$ and $x_t\in \cX$,
\begin{multline}\label{eq:dot_x_t_z}
\psi(x_{t})+ \frac1{\deltat}\left< z, \lambda(x_{t})\right> - \frac{\|z\|^2}{2 \deltat} + \log |\det Q|  \\ = \psi_*(x_{t}) + \frac1{\deltat}\left<Q z + b, \lambda_*(x_{t})\right> - \frac{\|Q z + b\|^2}{2\deltat}.
\end{multline}
Taking derivatives with respect to $z$, we get 
\begin{equation}\label{eq:dot_x_t_z_der}
\lambda(x_{t}) - z = Q^\top\lambda_*(x_{t}) - Q^\top Q z - Q^\top b.
\end{equation}
Taking derivatives again we get $Q^\top Q = I$, thus $Q$ is orthogonal. Replacing this back in \eqref{eq:dot_x_t_z_der}, we get
\begin{equation}\label{eq:dot_x_t_z_der_simplified}
\lambda_*(x_{t}) = Q\lambda(x_{t}) + b.
\end{equation}
Replacing \eqref{eq:lambda_def}, letting $f(z)=x_t$ and using \eqref{eq:ftildefcorrespondence}, we get for all $z\in \cZ$,
\begin{equation}
Qz + b - \mu_*(Qz + b)\deltat = Qz - Q\mu(z)\deltat + b,
\end{equation}
which implies $\mu(z) = Q^\top\mu_*(Q z + b)$ for all $z\in \mathcal{Z}$. 
Finally, replacing all equations obtained in \eqref{eq:dot_x_t_z}, and using \eqref{eq:psi_def} and $|\det Q| = 1$ (since $Q$ is orthogonal), we get
\begin{equation}
\log p_\gamma(f^{-1}(x_{t})) = \log p_{\gamma_*}(f_*^{-1}(x_{t})),
\end{equation}
and \eqref{eq:ident_p_gamma} follows from taking $x_t = f(z)$ and using \eqref{eq:ftildefcorrespondence}.
\qed
\vspace{-10pt}
\section{Supplementary results regarding practical considerations}\sectionlabel{sec:supplementary}

Here we complement present the proof of several results presented in the main submission.

\subsection{Proof of Theorem~\ref{thm:time_lamperti}}
\begin{proof}
We prove the lemma by constructing the functions $g$ and $\tilde \mu$ such that the SDE that governs $Y_t = g(Z_t, t)$ is given by \eqref{eq:general_Y_t}. By Ito's lemma \cite{kunita1967square}, we have
\begin{align}\label{eq:pre_Z_t}
	\nonumber dY_t & = \left(\frac{\partial g}{\partial t}(Z_t, t) + \frac{\partial g}{\partial y}(Z_t, t)\tilde \mu(Z_t, t) + \frac12 \Delta g(Z_t, t) \right) \d t \\ & + \frac{\partial g}{\partial y}(Z_t, t) \d W_t,
\end{align}
where $\frac{\partial g}{\partial y}(z , t)$ is the Jacobian of $g$, only in terms of $y$, and $\Delta g$ is the Laplacian, defined by
\begin{equation*}
	(\Delta g(z, t))_i = \sum_{k=1}^d \frac{\partial^2 g(z, t)_i}{\partial y_k^2}.
\end{equation*}
We now choose $h$ and $g$ as in Lemma \ref{lemma:h_lemma}, noting that $h$ is the inverse of $g$, therefore $Z_t = h(Y_t,t)$ and
\begin{align*}
	\frac{\partial g}{\partial y}(Z_t, t)
	&= \frac{\partial g}{\partial y}(h(Y_t, t), t),\\
	&= \sigma(Y_t,t).
\end{align*}
We just obtained that the diffusion terms in \eqref{eq:general_Y_t} and \eqref{eq:pre_Z_t} are equal, and the drift terms coincide if we define
\begin{equation*}
	\tilde \mu(z, t) := \frac{\partial g}{\partial y}(z, t)^{-1}\left(\mu(g(z,t), t) - \frac{\partial g}{\partial t}(z, t) -  \frac12 \Delta g(z, t) \right).
\end{equation*}
\end{proof}

\begin{lemma}\label{lemma:h_lemma}
	Suppose $\sigma$ follows conditions (i) and (ii) in Theorem \ref{thm:canonical_sigma_} and define
	\begin{equation}\label{eq:h_def}
		h(y,t):=\int_0^1 \sigma(\tau y,t)^{-1} y\d\tau.
	\end{equation}
	Then the following conditions hold:
	\begin{enumerate}[(i)]
		\item \begin{equation}
			\frac{\partial h}{\partial y}(y,t) = \sigma(y,t)^{-1}.
		\end{equation}
		\item There exists a function $g:\R^{d}\times\T \to \R^d$ such that for all $z\in \R^d, t \in \T$, $g(z,t) = y$ whenever $h(y,t)=z$. 
		\item $g$ is differentiable everywhere and
		\begin{equation*}
			\frac{\partial g}{\partial y}(h(y,t), t) = \sigma(y,t).
		\end{equation*}
	\end{enumerate}
\end{lemma}
\begin{proof}
The proof is largely algebraic manipulations, so we leave the full details to the supplementary materials. 
\end{proof}
\vspace{-5pt}
\subsection{Interpretability for learnable diffusion coefficients}\label{app:Intlearndiffcoeff}
Here we present a result similar to Theorem \ref{thm:identifiability_main}, but considering a learnable diffusion coefficient. That is, we rewrite \eqref{eq:Zt_pSDE},
\begin{align*}\label{eq:Zt_pSDE_diffusion}
&p_{\mu,\sigma} (Z_{t+\deltat}|Z_t, t) = \frac1{(2\pi \deltat)^\frac{d}2\det \sigma(Z_t,t)^\frac12}\\&\exp\left(-\frac1{2\Delta t} (Z_{t+\deltat} - \lambda(Z_t))^\top \sigma(Z_t,t)^{-1} (Z_{t+\deltat} - \lambda(Z_t))\right).
\end{align*}
where $\sigma(Z_t, t)$ is the diffusion coefficient and $\lambda(Z_t)=Z_t + \deltat\mu(Z_t,t)$. We first state the result, then explain some of our reservations against it, and finally present its proof.

A brief note on notation: for a symmetric $n \times n$ matrix $M$, we denote by $\vecsym{M}$, a vector of dimension $\binom{n+1}{2}$, which consists of $M$ flattened to a vector, such that the entries off-diagonal only appear once, and are multiplied by $\sqrt 2$. This definition implies that, for 2 symmetric matrices $A$ and $B$, we have $\tr(AB) = (\vecsym{A})^\top  \vecsym{B}$.
Moreover, we denote by $u\oplus v$ the concatenation of vectors $u$ and $v$.
\begin{theorem}\label{thm:identifiability_arbitrary}
	Suppose that the true generative model with arbitrary diffusion coefficients has parameters $\theta^*= (f^*, \mu^*, \sigma^*, \gamma^*)$, and that the following conditions hold:
	\begin{enumerate}
		\item The set $\{x\in \mathcal X|\varphi_\epsilon(x)=0\}$ has measure zero, where $\varphi_\epsilon$ is the characteristic function of the density $p_\epsilon$ defined in \eqref{eq:Xt_epsilon}.
		\item $f^*$ is injective and differentiable.
		\item Letting $N=\binom{n+1}{2}+n$, there exist $N+1$ vectors $z_0, \dots, z_N$ and scalars $t_0, \dots, t_N$ such that the vectors $\bm{\Lambda}_1-\bm{\Lambda}_0, \dots, \bm{\Lambda}_N-\bm{\Lambda}_0$, with
		\begin{equation}
		\hspace{-10pt}
			\bm{\Lambda}_i := \vecsym(\sigma(z_i,t_i)^{-1})\oplus \left(-2\sigma(z_i,t_i)^{-1} (z_i -\mu(z_i,t_i))\right),
		\end{equation}
		are linearly independent.
		
	\end{enumerate}
	Then if $\theta = (f, \mu, \sigma, \gamma)$ are other parameters that yield the same generative distribution, that is
	\begin{equation}
	p_{\theta}(x_{t+\deltat}, x_t) = p_{\theta^*}(x_{t+\deltat}, x_t)\quad \forall x_{t+\deltat}, x_t \in \R^n,
	\end{equation}
	then $\theta$ and $\theta^*$ are equal up to an affine transformation. That is, there exists an invertible matrix $A$ and a vector $b$, such that for all $z\in \R^d$:
	\begin{equation}
	f(z) = f^*(A z + b),
	\end{equation}
	\begin{equation}
	\mu(z, t) = A^{-1} \mu^*(A z + b, t)\quad \forall t\in \T,
	\end{equation}
	\begin{equation}
	\sigma(z, t) = A^{-1} \sigma^*(A z + b, t) A^{-\top}\quad \forall t\in \T,
	\end{equation}
	and
	\begin{equation}\label{eq:ident_p_gamma_ld}
	p_{\gamma}(z) = |\det A|^{-1} p_{\gamma^*}(A z + b).
	\end{equation}
\end{theorem}

Before we show the details of the proof, we explain why we have chosen not to include this result in the main submission. The main reason is that condition 3, in theorem statement, is not satisfied for simpler diffusion coefficients, such as a constant diffusion coefficient. We felt our theory was not satisfactory if it did not apply for a simple Brownian motion, which is one of the most simple SDEs that exist.

Other issue is related to identifiability: first we lose uniqueness up to an isometry, getting an affine transformation, and then we lose the connection between SDEs arising from the It\^o's lemma. It\^o's lemma implies that employing a change of variable effectively leads to another SDE, and therefore we can never estimate the true latent variable up to this change of variables. However Theorem \ref{thm:identifiability_arbitrary} fails to capture this.
We finally present the proof of this Theorem in the supplementary materials due to its similarity to the proof of Theorem~\ref{thm:identifiability_main}.
\subsection{Proof of Theorem~\ref{thm:latent_size}}\sectionlabel{sec:latent_proof}
\begin{proof}
For this proof, since it concerns recovering the latent dimension size, we assume that $\hat{A}, \mu, \sigma$ are well recovered in the sense that \eqref{eq:L_def} is minimized given the conditions on the latent size in $\hat{A}$.
Applying It\^o's lemma, we obtain a new SDE for the transformed data. 
Since the transformation is linear, it is easy to characterize the distribution of the transformed space.
Using the Euler-Maruyama discretization as above, we obtain a distribution on the increments $I_t = X_{t+ \Delta t} - X_{t}$ of $X_t$
\begin{align}
   & I_t  \sim \mathcal{N}\left(A\mu(A^{-1} X_t,t)\Delta t, A\sigma(A^{-1} X_t, t)A^\top \Delta t \right) \\
   \nonumber & Z_{t+ \Delta t} - Z_{t} \sim \mathcal{N}\left(\mu(Z_t,t)\Delta t, \sigma(Z_t, t)\Delta t\right)
    \label{eq:linear_likelihood}
\end{align}
and recover the likelihood of $\mathbf{X}$
\begin{align*}
    & \log \mathcal{L}( X_{t+\Delta t} | X_t , A)  = -\frac12 \log \det A\sigma(A^{-1} X_t, t)A^\top \Delta t \\  &- \frac12 \left( I_t - A\mu(A^{-1} X_t, t) \Delta t \right ) \left (A\sigma(A^{-1} X_t, t)A^\top  \Delta t \right )^{-1} \\ & \hspace{20pt} \left( I_t - A \mu(A^{-1} X_t, t) \Delta t \right )^\top   - \frac d2 \log 2\pi 
\end{align*}
where determinants and inverses are understood as pseudodeterminants and pseudoinverses. 
To recover the size of the latent dimension, we must estimate the map $\hat{A} \in \mathbb{R}^{n \times j}$ from a $j$ dimensional latent space.
We then consider the cases when $j> d$ and $j < d$ to show that at the minimum of \eqref{eq:aic}, $j=d$.

$(j > d)$: Suppose $j > d$, then it is straightforward to show that $\hat{A}^\top \hat{A}$ is singular since at minimum $\rank \hat{A} \leq d$.
When $\hat{A}^\top \hat{A}$ is singular, $\mathcal{L}(\hat{A})$ is undefined since $\left ( \hat{A} \sigma \hat{A}^\top  \right )^{-1}$ and $\hat{A}^{-1}$ do not exist.
Followingly, \eqref{eq:aic} will be undefined, therefore when $j > d$, the undefined likelihood makes $j$ the incorrect choice.

$(j < d)$: Suppose now that $j < d$. 
Since $\rank \sigma(x,t) = d$, then $ \rank A\sigma(Z_t, t)A^\top \Delta t = d$ but  $\rank \hat{A}\sigma(Z_t, t)\hat{A}^\top \Delta t = j < d$.
Since $\rank \Cov(I_t) = d$ but the estimated rank of the covariance is $j$, $\hat{A}$ cannot achieve the maximum likelihood estimate. 
Then $\mathcal{L}(\hat{A}) < \mathcal{L}(A)$.

Finally, since $j \nless d$ and $j \ngtr d$, $j = d$.
\end{proof}
\vspace{-15pt}
\section{Calculating optimal orthogonal and affine transformations}\sectionlabel{sec:orthogonal}
Here we describe the minimizers of \eqref{eq:orthog_1}. Let $A$ be a $d\times N$ matrix with columns given by $A_t = \tilde f^{-1}(X_t)$, and let $B$ a $d\times N$ matrix with columns given by $B_t = Z_t$. Then we have \eqref{eq:orthog_1} is equivalent to:
\begin{equation}
		\mathcal{L}_\text{latent} = \frac{1}{N}\min_{Q,b} \| Q A + b \bm{1}^\top   - B \|_F^2, 
	\label{eq:AB_form}
\end{equation}
where $\bm{1}$ is the all-ones vector and $\|\cdot\|_F$ denotes Frobenious norm, defined by $\|M\|_F=\sqrt{\tr(M^\top  M)}$. Followingly, we use the Frobenious dot-product of matrices, defined as $\left<M,O\right>=\tr(M^\top  O)$. We present closed form solutions of
\begin{equation}
\begin{aligned}
\min\,&\| Q A + b \bm{1}^\top   - B \|_F^2\\
\text{s.t.}\,\,& Q\in \mathcal{Q}, b\in \R^d
\end{aligned}
\label{eq:D.2}
\end{equation}
for both cases when $\mathcal{Q}$ is the space of square matrices and of orthogonal matrices. We first calculate $b$ in terms of $Q$.
\begin{align*}
& \| Q A + b \bm{1}^\top   - B \|_F^2 \\ &= \| Q A - B \|_F^2 + \left<b \bm{1}^\top ,  b \bm{1}^\top  - 2 (Q A - B) \right>\\
&= \| Q A - B \|_F^2 + \tr\left(\bm{1} b^\top  (b \bm{1}^\top  - 2 (Q A - B))\right)\\
&= \| Q A - B \|_F^2 + \tr\left(N b^\top  b - 2 b^\top  (Q A - B))\bm{1}\right)
\end{align*}
Since this is a convex function of $b$, the minimum is obtained when we set the gradient to zero
\begin{align}
\nonumber &2N b - 2 (Q A - B)\bm{1} = \bm{0}\\
\Rightarrow\,& b = \frac1N (Q A - B) \bm{1} \label{eq:b_def}
\end{align}
Replacing $b$ in \eqref{eq:D.2}, we get
\begin{equation}
		\min_{Q}\| Q \tilde A - \tilde B \|_F^2,
	\label{eq:D.3}
\end{equation}
where $\tilde A = A - \frac1N A\bm{1}\bm{1}^\top $ and $\tilde B = B - \frac1N B\bm{1}\bm{1}^\top $. We note that $\frac1N A\bm{1}$ and $\frac1N B\bm{1}$ are the average of the columns of $A$ and $B$, respectively, thus $\tilde A$ and $\tilde B$ are centered versions of $A$ and $B$.
We first find the minimum of \eqref{eq:D.3} for general square matrices. In this case the objective is again a convex function of the entries of $Q$, so we can find the minimizer by equating the gradient of $Q$.
\begin{align*}
&2Q \tilde A \tilde A^\top  - 2 \tilde B \tilde A^\top  = \bm{0}\\
\Rightarrow\,&Q = \tilde B \tilde A^\top  (\tilde A \tilde A^\top )^{-1}
\end{align*}

On the other hand, if $Q$ is orthogonal, then \eqref{eq:D.3} is an instance of the orthogonal Procrustes problem \cite{gower2004procrustes}, and letting $USV^\top  = \tilde B \tilde A^\top $ be the singular value decomposition of $\tilde B \tilde A^\top $, the minimizer is $Q=UV^\top $. 
With $Q$ calculated, we replace it in \eqref{eq:b_def} to finally calculate $b$.
\section{Cramer-Rao bounds for estimating the drift coefficient}\sectionlabel{sec:crlb}

Here we derive the Cram\'er-Rao lower bound (CRLB) we use for estimating SDEs. The CRLB gives an information theoretically lower bound on the MSE of any estimator, and in particular it is also a lower bound for estimating $\mu$ using the proposed VAE. However, it is hard to calculate the CRLB for some of the SDEs considered, thus we focus on the simplified problem: determining $\mu$ up to a global shift. That is,
$$\mu(z)= \mu^*(z) + \eta,\quad \forall z\in \cZ,$$
where the function $\mu^*(z)$ is known but $\eta\in \R^d$ is unknown. We note that this is the exact CRLB for the constant drift SDE, and is still a lower bound on the estimation error for the OU process.

Using the Euler-Maruyama approximation, the increments of the SDE are distributed as $\mathcal{N}((\mu^*(z) + \eta)\Delta t, \sigma \Delta t)$. The CRLB for an estimator of the mean $\hat \mu$ of the Gaussian distribution $\mathcal{N}(\mu, \sigma)$ is
$$
\Cov(\hat \mu) \succeq \frac1N \sigma,
$$
which implies that $\E[\|\hat \mu - \mu\|^2]\ge \frac1N \tr\sigma$. Substituting the values from our increment distribution, and taking into account that $\sigma=I$, we get
\begin{align*}
    \mathbb{E}[\|(\hat \eta -\eta)\Delta t\|^2] 
    & \geq  \frac{\tr(I) \Delta t}{N} \\
    \mathbb{E}[\|\hat \mu(z) - \mu(z) \|^2] & \ge  \frac{d}{\Delta t N}
\end{align*}
\vspace{-5pt}

\section*{Acknowledgment}
This work was supported in part by Office of Naval Research Grant  N00014-18-1-2244. 
AH was supported by the National Science Foundation Graduate Research Fellowship.
The authors would also like to thank Jessica Loo and Joe Kileel for helpful feedback on the paper. 


\ifCLASSOPTIONcaptionsoff
  \newpage
\fi


\bibliographystyle{IEEEtran}
%
\bibliography{IEEEabrv,ref}

%








\clearpage

\section{Supplemental Proofs}
\subsection{Proof of Lemma~\ref{lemma:h_lemma}}

\begin{proof}
	First, \eqref{eq:no_curl_cond} implies that for all $j,k$
	\begin{align*}
		\frac{\partial}{\partial y_k} \left(\sigma(y, t)^{-1}\right) e_j 
		&= - \sigma(y, t)^{-1} \frac{\partial\sigma(y, t)}{\partial y_k} \sigma(y, t)^{-1} e_j  \\
		&= - \sigma(y, t)^{-1} \frac{\partial\sigma(y, t)}{\partial y_j} \sigma(y, t)^{-1} e_k, \\
		&= \frac{\partial}{\partial y_j} \left(\sigma(y, t)^{-1}\right)  e_k,
	\end{align*}
	and (i) follows from \cite[Theorem 11.49]{lee2013smooth}. We now prove that $h(\cdot,t)$ is an one-to-one function, thus proving its inverse $g(\cdot, t)$ exists and is well-defined.  
	
	We start by showing that for all $t$ the set $\mathcal{H}_M:=\{y:\|h(y,t)\|\le M\}$ is compact. Since $h$ is continuous, $\mathcal{H}_M$ is a closed set, thus it remains to prove it is bounded. Condition (i) in Theorem \ref{thm:canonical_sigma_} implies not only that $\sigma(y, t)$ is positive definite but also that its norm is bounded by $\|y\|$.
	\begin{align}
		\|\sigma(y, t)\| 
		&\le \|\sigma(y, t)-\sigma(0, t)\| + \|\sigma(0,t)\|,\\
		&\le D \|y\| + \|\sigma(0,t)\|,\\
		&\le \tilde D (1+\|y\|),
	\end{align}
	where $\tilde D = \max\{D, \max_{t\in \T} \|\sigma(0,t)\|\}$. This implies the eigenvalues of $\sigma(y, t)$ are upper-bounded by $(1+ \|y\|)\tilde D$, and since $\sigma(y, t)$ is a PSD matrix, the eigenvalues of its inverse are lower bounded by $1/((1+ \|y\|)\tilde D)$, thus
	\begin{equation}\label{eq:Sigma1_PSD_lowerbound}
		w^T \sigma(y, t)^{-1} w \ge \frac{\|w\|^2 }{(1+ \|y\|)\tilde D}\quad \text{for all} \quad y,w\in\R^d.
	\end{equation}
	Applying this and the Cauchy-Schwarz inequality,
	\begin{align*}
		\|y\|\|h(y,t)\| &\ge y^T h(y,t)\\
		&=\int_0^1 y^T \sigma(\tau y,t)^{-1} y\d\tau,\\
		&\ge \int_0^1 \frac{\|y\|^2 }{(1+  \tau \|y\|)\tilde D}\d\tau,\\
		&= \frac{\|y\|}{\tilde D}\log\left(1+\|y\|\right).
	\end{align*}
	Therefore, if $M\ge \|h(y,t)\|$, then $M\ge \frac{1}{\tilde D}\log\left(1+\|y\|\right)$, thus
	\begin{equation*}
		\|y\| \le \exp(M \tilde D) - 1,
	\end{equation*}
	which proves $\mathcal{H}_M$ is bounded. We now prove $h$ is surjective.
	Let $z\in \R^d$ and
	\begin{equation}\label{eq:surjectopteq}
		y_* = \argmin_{y\in \mathcal{H}_{1 + 2 \|z\|}} \|h(y,t) - z\|^2.
	\end{equation}
	Since $\mathcal{H}_{1 + 2 \|z\|}$ is compact, the minimum is achieved by a point in $\mathcal{H}_{1 + 2 \|z\|}$, and this point is not in the boundary. Since by definition $h(0, t) = 0$, we have $0\in \mathcal{H}_{1 + 2 \|z\|}$ and for any element $w\in \partial\mathcal{H}_{1 + 2 \|z\|}$
	\begin{align*}
		\|h(w,t) - z\| &\ge \|h(w,t)\| - \|z\|, \\
		&= 2\|z\| + 1 - \|z\|, \\
		&> \|z\| = \|h(0, t) - z\|,
	\end{align*}
	thus $w$ does not achieve the minimum. Since the minimum is achieved by an interior point of $\mathcal{H}_{1 + 2 \|z\|}$ and $h$ is differentiable, $y_*$ is a critical point of \eqref{eq:surjectopteq}. We then have
	\begin{align*}
		0 &= \nabla\|h(y_*,t) - z\|^2,\\
		&=2\frac{\partial h}{\partial y}(y_*,t) (h(y_*,t) - z),\\
		&=2\sigma(y_*,t)^{-1} (h(y_*,t) - z).
	\end{align*}
	Since $\sigma(y_*,t)^{-1}$ is non-singular, we must have $h(y_*,t) = z$, thus $h$ is surjective. We now prove that $h(y,t)\neq h(w,t)$ for all $y,w \in \R^d$. By the Fundamental Theorem of Calculus
	\begin{align*}
		(w-y)^T &\left(h(w,t)-h(y,t)\right) 
		\\&= \int_0^1 (w-y)^T \frac{\partial h}{\partial y} (y + \tau(w-y),t) (w-y)\d\tau,\\
		&= \int_0^1 (w-y)^T \sigma(y + \tau(w-y),t)^{-1} (w-y)\d\tau
		\\ &> 0,
	\end{align*}
	where the last line follows from \eqref{eq:Sigma1_PSD_lowerbound}, thus $h(y,t)\neq h(w,t)$, $h$ is bijective, and $g$ in (ii) is well defined. Finally, (iii) follows from the Inverse Function Theorem
	\begin{align*}
		\frac{\partial g}{\partial y}(h(y,t), t)&= \frac{\partial h}{\partial y}(y, t)^{-1}, \\ & = \sigma(y, t).
	\end{align*}
\end{proof}	

\subsection{Proof of Theorem \ref{thm:identifiability_arbitrary}}
\begin{proof}
The proof is again very similar to the proof of Theorem \ref{thm:identifiability_main}. The 3 steps are the same as in the previous proof, and step (I) follows exactly in the same way, so we start on step (II).

\subsubsection*{Step (II)}
From Step (I), we have that for all $x_t, x_{t + \Delta t}$ in $\cX$,
\begin{align}
\nonumber & \log\vol J_{f_*^{-\!1}}(x_{t + \Delta t}) + \log\vol J_{f_*^{-\!1}}(x_{t}) \\ \nonumber & + \log p_{\mu_*,\sigma_*}(f_*^{-1}(x_{t + \Delta t})|f_*^{-1}(x_{t}), t) +  \log p_{\gamma_*}(f_*^{-1}(x_{t}))
\\ \nonumber & \hspace{-1pt} = \log\vol J_{f^{-\!1}}(x_{t + \Delta t}) + \log\vol J_{f^{-\!1}}(x_{t}) \\& + \log p_{\mu,\sigma}(f^{-1}(x_{t + \Delta t})|f^{-1}(x_{t}), t) +  \log p_\gamma(f^{-1}(x_{t})).
\label{eq:noiseless_p_equality_ld}
\end{align}
Let
\begin{equation}\label{eq:lambda_def_ld}
\lambda(x_{t})= f^{-1}(x_{t}) - \mu(f^{-1}(x_{t}), t)\deltat,
\end{equation}
\begin{equation}\label{eq:eta_def_ld}
\eta(x_{t})= \sigma(f^{-1}(x_{t}), t)^{-1} \lambda(x_{t}),
\end{equation}
and define analogously $\lambda_*(x_{t}),\eta_*(x_{t})$. Let $z_t=f^{-1}(x_{t})$, and similarly for $z_{t+\deltat}$. We then have
\begin{align*}
& \log p_{\mu,\sigma}(z_{t + \Delta t}|z_{t}, t) \\
&= -\frac12\log\det \sigma(z_{t}, t)\\ & - \frac1{2\Delta t} (z_{t+\deltat} - \lambda(x_t))^T\sigma(z_t,t)^{-1} (z_{t+\deltat} - \lambda(x_t)),\\
&=\mathscalebox{.95}{-\frac12\log\det \sigma(z_{t}, t)}\\
&\hspace{3pt} \mathscalebox{.95}{- \frac1{2\Delta t}\left(z_{t+\deltat}^T\sigma(z_t,t)^{-1}z_{t+\deltat} - 2 z_{t+\deltat}^T \eta(x_{t}) + \lambda(x_{t})^T\eta(x_{t})\right).}
\end{align*}
We now write this as an augmented linear system. Let $\mathcal{A}(z)=\vecsym(zz^T)\oplus z$,
\begin{equation}
\xi(x_t)=\vecsym(\sigma(f^{-1}(x_t),t)^{-1})\oplus (-2\eta(x_{t})),
\end{equation}
and define $\xi_*(x_t)$ analogously.
We have
\begin{multline*}
z_{t+\deltat}^T\sigma(z_t,t)^{-1}z_{t+\deltat}\\ =\left<\vecsym(z_{t+\deltat}z_{t+\deltat}^T), \vecsym(\sigma(z_t,t)^{-1})\right>,
\end{multline*}
thus
\begin{align*}
& \log p_{\mu,\sigma}(z_{t + \Delta t}|z_{t}, t) \\
&= -\frac12\log\det \sigma(z_{t}, t) \\ & - \frac1{2\Delta t}\left(\left<\mathcal{A}(z_{t+\deltat}), \xi(x_t)\right> +\lambda(x_{t})^T\eta(x_{t})\right),\\
&= -\frac12\log\det \sigma(f^{-1}(x_t), t)\\ & - \frac1{2\Delta t}\left(\left<\mathcal{A}(f^{-1}(x_{t+\deltat})), \xi(x_t)\right> +\lambda(x_{t})^T\eta(x_{t})\right).
\end{align*}
Finally define
\begin{align}\label{eq:psi_def_ld}
\nonumber \psi(x_{t}) &= \vol J_{f^{-1}}(x_{t}) + \log p_\gamma(f^{-1}(x_{t})) \\ &  -\frac12\log\det \sigma(f^{-1}(x_t), t) - \frac{\left<\lambda(x_{t}),\eta(x_{t})\right>}{2 \deltat},
\end{align}
\begin{equation}\label{eq:varphi_def_ld}
\varphi(x_{t + \Delta t}) = \vol J_{f^{-1}}(x_{t + \Delta t}),
\end{equation}
and $\psi_*$ and $\varphi_*$ analogously. We finally have for all $x_t, x_{t + \Delta t}$ in $\cX$, an equation similar to \eqref{eq:dot_x_t_canonical}.
\begin{align}\label{eq:dot_x_t_canonical_ld}
\nonumber & \psi(x_{t}) - \frac1{2\deltat}\left< \mathcal{A}(f^{-1}(x_{t + \Delta t})), \xi(x_{t})\right> + \varphi(x_{t + \Delta t}) \\ &= \psi_*(x_{t}) - \frac1{2\deltat}\left<\mathcal{A}(f_*^{-1}(x_{t + \Delta t})), \xi_*(x_{t})\right> + \varphi_*(x_{t + \Delta t}).
\end{align}
We now proceed in the same way as before to obtain a similar equality for all $x_{t+\Delta t} \in \cX$.
\begin{equation}\label{eq:augmented_linear_dependence}
 \mathcal{A} (f_*^{-1}(x_{t + \Delta t})) = {\mathbf{\Xi}}_*^{-T}{\mathbf{\Xi}}^T \mathcal{A} (f^{-1}(x_{t + \Delta t})) + {\mathbf{\Xi}}_*^{-T}\bm{\psi},
\end{equation}
with the exception that the invertibility of ${\mathbf{\Xi}}_*^T$ is now a consequence of condition 3 in the theorem statement. To prove ${\mathbf{\Xi}}_*^{-T}{\mathbf{\Xi}}$ is invertible, let $z_0, z_1, \dots, z_N\in \R^d$, with $N$ defined as in the theorem statement, such that the vectors $\mathcal{A}(z_i)-\mathcal{A}(z_0)$, $i=1,\dots,N$ are linearly independent. We note that the entries of $\mathcal{A}(z)$ are linear independent polynomials so finding such vectors is always possible. Let $x_i=f_*(z_i),$ $i=0,\dots,N$, define
$\mathbf{A}_*$ as the $N\times N$ matrix with columns $\mathbf{A}_{*,i}$
\begin{equation*}
\mathbf{A}_{*,i}=\mathcal{A} (f_*^{-1}(x_{i}))-\mathcal{A} (f_*^{-1}(x_{0})),
\end{equation*}
and define $\mathbf{A}$ analogously. Equation \eqref{eq:augmented_linear_dependence} implies
\begin{equation}\label{eq:augmented_linear_dependence_invertible}
\mathbf{A}_* = {\mathbf{\Xi}}_*^{-T}{\mathbf{\Xi}}^T \mathbf{A}.
\end{equation}
Since $\mathbf{A}_*$ is invertible by construction, so is ${\mathbf{\Xi}}_*^{-T}{\mathbf{\Xi}}^T$. From \eqref{eq:augmented_linear_dependence}, we can now get a linear dependence between $f_*^{-1}$ and $f^{-1}$. Let $\nu(z) = f_*^{-1}(f(z))$, then
\eqref{eq:augmented_linear_dependence} implies that there are symmetric matrices $M_i$, vectors $v_i$ and scalars $c_i$ such that
\begin{equation}\label{eq:nu_1}
\nu(z)_i = z^T M_i z + v_i^T z + c_i,
\end{equation}
and that there are symmetric matrices $\tilde M_i$, vectors $\tilde v_i$ and scalars $\tilde c_i$ such that
\begin{equation}\label{eq:nu_2}
\nu(z)_i^2 = z^T \tilde M_i z + \tilde v_i^T z + \tilde c_i.
\end{equation}
Subtracting the square of \eqref{eq:nu_1} with the \eqref{eq:nu_2} for all $z$ implies that $M_i= 0$, $\tilde M_i = v_i v_i^T$, $\tilde v_i = 2 c_i v_i$ and $\tilde c_i = c_i^2$. This and the invertibility of ${\mathbf{\Xi}}_*^{-T}{\mathbf{\Xi}}^T$ finally imply that there is an invertible matrix $A$ and a vector $b$ such that for all $x_{t+\deltat}$
\begin{equation}\label{eq:ftildefcorrespondence_ld}
f_*^{-1}(x_{t + \Delta t}) = A f^{-1}(x_{t + \Delta t}) + b.
\end{equation}
\subsubsection*{Step (III)}
Equation \eqref{eq:ftildefcorrespondence_ld} implies that for all $z\in \cZ$
\begin{align*}
f(z)&=f_*(f_*^{-1}(f(z))),\\
&=f_*(A f^{-1}(f(z)) + b),\\
&=f_*(A z + b).
\end{align*}
Moreover $J_{f}=J_{f_*} A	$, and $\log\vol J_{f^{-1}} = \log\vol J_{f_*^{-1}} + \log|\det A|$. Now replacing $x_{t + \Delta t}$ by $f(z)$ on \eqref{eq:dot_x_t_canonical_ld} and using \eqref{eq:ftildefcorrespondence_ld} and \eqref{eq:psi_def_ld}, we get for all $z\in \cZ$ and $x_t\in \cX$
\begin{multline}\label{eq:dot_x_t_z_ld}
\psi(x_{t}) - \frac1{\deltat}\left< z, \eta(x_{t}))\right> + \frac{z^T \sigma(f^{-1}(x_t),t)^{-1} z}{2 \deltat}  + \log |\det A| \\
= \psi_*(x_{t}) - \frac1{\deltat}\left<A z + b, \eta_*(x_{t})\right> \\  + \frac{(A z + b)^T\sigma_*(f_*^{-1}(x_t),t)^{-1}(A z + b)}{2\deltat}
\end{multline}
Taking derivatives with respect to $z$, we get 
\begin{multline}\label{eq:dot_x_t_z_der_ld}
-\eta(x_{t}) + \sigma(f^{-1}(x_t),t)^{-1} z \\ = -A^T\eta_*(x_{t}) + A^T\sigma_*(f_*^{-1}(x_t),t)^{-1}(A z + b)
\end{multline}
Taking derivatives again we get
\begin{equation}\label{eq:sigma_ld}
\sigma(f^{-1}(x_t),t)^{-1} = A^T\sigma_*(f_*^{-1}(x_t),t)^{-1}A 
\end{equation}
Letting $f(z)=x_t$ and using \eqref{eq:ftildefcorrespondence_ld}, we get for all $z\in \cZ$,
\begin{equation}\label{eq:sigma__ld}
\sigma(z,t) = A^{-1}\sigma_*(Az + b,t)A^{-T}
\end{equation}
Replacing this back in \eqref{eq:dot_x_t_z_der_ld}, using \eqref{eq:eta_def_ld} and multiplying by the inverse of \eqref{eq:sigma_ld}, we get
\begin{equation}\label{eq:dot_x_t_z_der_simplified_ld}
\lambda(x_{t}) = A^{-1}\lambda_*(x_{t}) - A^{-1} b,
\end{equation}
or $\lambda_*(x_{t})=A\lambda(x_{t}) + b$. Using \eqref{eq:sigma_ld}, \eqref{eq:lambda_def_ld}, letting $f(z)=x_t$ and using \eqref{eq:ftildefcorrespondence}, we get for all $z\in \cZ$,
\begin{equation}
Az + b - \mu_*(Az + b, t)\deltat = Az - A\mu(z,t)\deltat + b
\end{equation}
which implies $\mu(z, t) = A^{-1}\mu_*(A z + b,t)$ for all $z\in \mathcal{Z}$. 
Finally, replacing all equations obtained in \eqref{eq:dot_x_t_z_ld}, and using \eqref{eq:psi_def_ld}, we get
\begin{equation}
\log p_\gamma(f^{-1}(x_{t})) = \log p_{\gamma_*}(f_*^{-1}(x_{t})) - \log |\det A|,
\end{equation}
and \eqref{eq:ident_p_gamma_ld} follows from taking $x_t = f(z)$ and using \eqref{eq:ftildefcorrespondence_ld}.
\end{proof}

\section{Hyperparameters}
\begin{table*}[t]
    \centering
    \begin{footnotesize}
    \begin{tabular}{@{}lccccc@{}}
    \toprule
        Dataset & Balls & Digits & Wasserstein & Balls + Wasserstein & DNA \\ \hline
        \multirow{2}{*}{Batch Size} & \multicolumn{5}{c}{Full Trajectory} \\
          & (800) & (100) & (800) & (800) & (100) \\ 
        Validation Size & \multicolumn{5}{c}{100} \\ 
        AE LR & \multicolumn{4}{c}{0.001} & 0.0001 \\
        $\hat{\mu}$ LR & \multicolumn{4}{c}{0.001} & 0.0001 \\
        $\hat{\mu}$ Width & \multicolumn{5}{c}{16}  \\
        $\hat{\mu}$ Depth & \multicolumn{5}{c}{4}  \\
        $\hat{\mu}$ Activation & \multicolumn{5}{c}{Softplus}  \\
        $\tau$ & 0.01 & 0.005 & 0.05 & 0.01 & 0.01 \\
        $\nu$ & \multicolumn{5}{c}{0} \\
        Optimizer & \multicolumn{5}{c}{Adam} \\
        \multirow{2}{*}{LR Decay} & \multicolumn{5}{c}{Exponential} \\
        & 0.997 & 0.997 & 0.999 & 0.999 & 0.9998 \\
        Epochs & \multicolumn{3}{c}{1500} & 2000 & 150 \\ \bottomrule
    \end{tabular}
    \end{footnotesize}
    \caption{Hyperparameters for all experiments.}
    \label{tab:hyperparameters}
\end{table*}

\begin{table}
\centering
\begin{footnotesize}
\begin{tabular}[c]{@{}l@{}}
\toprule
\textbf{Encoder} \\  \midrule
Input size: (3, 64, 64)\\ 
Conv(3, 8, 5, 1, 2), LeakyReLU, MaxPool(2, 2, 0)\\ 
Conv(8, 16, 5, 1, 2), BN, LeakyReLU, MaxPool(2, 2, 0)\\ Conv(16, 32, 5, 1, 2), BN, LeakyReLU, MaxPool(2, 2, 0)\\ Conv(32, 64, 5, 1, 2), BN, LeakyReLU\\ 
Flatten \\ 
FC$_\mu$(4096, $d$), FC$_\sigma$(4096, $d$) \\ 
Output size : ($d$, ) \\
\bottomrule
\end{tabular}
\end{footnotesize}
\caption{Encoder architectures for all experiments. BN refers to batch normalization.}
\label{tab:arch_e}
\end{table}

\begin{table}[h!]
    \centering
    \begin{footnotesize}
    
\begin{tabular}[c]{@{}l@{}}
\toprule
\textbf{Decoder} \\ \midrule
Input size: ($d$, )\\
FC($d$, 4096)\\
Unflatten \\
U, Conv(64, 64, 5, 1, 2), BN, LeakyReLU\\ 
U, Conv(64, 32, 5, 1, 2), BN, LeakyReLU\\ 
U, Conv(32, 16, 5, 1, 2), BN, LeakyReLU\\ U, Conv(16, 8, 5, 1, 2), BN, LeakyReLU\\ 
U, Conv(8, 3, 5, 1, 2), Sigmoid \\ 
Output size : (3, 64, 64) \\
\bottomrule
\end{tabular} 
\end{footnotesize}
    \caption{Decoder architecture for all experiments. U refers to an upsampling layer.}
    \label{tab:arch_d}
\end{table}
The list of hyperparameters used for the experiments are available in Table~\ref{tab:hyperparameters}.
Architecture details for all experiments are given in Tables~\ref{tab:arch_e} and~\ref{tab:arch_d}.
For each run, we run the algorithm 3 times on the same run and choose the one with the lowest validation loss. 
We performed limited hyperparameter tuning, instead leaving most parameters the same across all datasets.
For the noise datasets, we use the same hyperparameters as without the noise. 

\section{Fluorescent DNA Details}
The original data are given as spatially varying raw counts over  period of 100 frames per video. 
Let $V \in \mathbb{N}^{100 \times 512 \times 512}$ represent a video with 100 frames with 512 pixels as the width and height. 
We define
$$
\bar{V} = V - \frac{1}{100} \sum_i V_{i,j,k}
$$
and 
$$
\tilde{V} = \frac{\bar{V} - \min \bar{V}}{\max \bar{V} - \min \bar{V}}.
$$
After normalization, we pass the frames through a $4\times 4$ Gaussian filter with $\sigma=3$ and compute a new image using a maximum filter over $3 \times 3$ blocks of the image.
From this, we compute the maximum value corresponding to the center of the molecule. 
We then take the average of all the centers as a guideline for a refinement iteration where we again compute the centers conditioned on the mean center. 

For the input to the neural network, we first compute 
$$
\hat{V} = \frac{V - \min V}{\max V - \min V}.
$$
and then apply the adaptive histogram equalization \cite{zuiderveld1994contrast} algorithm to each frame in the video. 

{
	\renewcommand\arraystretch{1.5}
	\begin{table*}[t!]
		\centering
		\captionsetup{width=.8\linewidth}
		\newcommand{\timesten}{\text{\footnotesize $\times10$}}
		\small
		\begin{tabular}{@{}lccccc@{}} \toprule
			Dataset & SDE Type & Oracle + \cite{duncker2019learning} & Proposed  & CRLB  & $p$-value \\ \midrule 
			
			\multirow{3}{*}{Balls} & Constant &  $2.37 (\pm 0.10) \timesten^{-1}$ & $5.07 (\pm  2.99) \timesten^{-1} $ & 	\multirow{3}{*}{$2.00 \timesten^{-1}$} & $0.0783$ \\ 
			
			& OU  &  $2.17(\pm 0.10) \timesten^{-1}$ & $9.20 (\pm 2.23) \timesten^{-1}$ & & $0.0001$ \\ 
			
			& Circle & $4.43(\pm 0.74) \timesten^{0}$ & $1.47 (\pm 0.62) \timesten^{0\phantom{-}}$ &  & $0.0001$ \\  
			Multiple Balls & OU & $2.74(\pm0.51)\timesten^{0}$ & $2.61(\pm 0.51) \timesten^{0}$ & $1.00 \timesten^{0\phantom{-}}$ & $0.6975$\\
			
			\cline{1-6}
			
			\multirow{3}{*}{Digits} & Constant & $1.92(\pm 0.49) \timesten^{-1}$ & $7.74(\pm 4.65) \timesten^{-1}$ &  	\multirow{3}{*}{$4.00 \timesten^{-1}$} & $0.0238 $ \\ 
			& OU &  $5.66(\pm 0.42) \timesten^{-1}$ & $7.69(\pm 4.91) \timesten^{-1}$ & & $0.3839 $ \\ 
			& Circle & $2.96 (\pm 0.22) \timesten^{0} $ & $2.15 (\pm 0.48) \timesten^{0\phantom{-}}$ & & $0.0089 $ \\ \cline{1-6}
			
			\multirow{3}{*}{Wasserstein} & OU & $1.54(\pm 0.48) \timesten^{-1}$  & $8.79(\pm 1.21) \timesten^{-1}$ & \multirow{3}{*}{$1.00 \timesten^{-1}$} & $<0.0001$ \\ 
			& Double Well & $5.37(\pm 2.45) \timesten^{0}$ & $1.45(\pm 0.76) \timesten^{0\phantom{-}}$ &  & $0.0091$ \\ 
			& GBM & $-$ & $1.23(\pm 2.45) \timesten^{-1}$  &  & $ -$ \\ \cline{1-6}
			
			\multirow{3}{*}{Ball + Wasserstein} & OU & $5.97(\pm 3.07) \timesten^{-1}$  & $7.08(\pm 2.79) \timesten^{-1}$ & \multirow{3}{*}{$3.00 \timesten^{-1}$}& $0.5661$  \\ 
			& Cauchy & $5.09(\pm 0.04) \timesten^{-1}$ & $8.82(\pm 7.36) \timesten^{-1}$ & & $0.2899$\\ 
			& Anisotropic & - &$1.74(\pm0.50)\timesten^{0\phantom{-}}$ & & - \\ 
			\bottomrule
		\end{tabular}
		\caption{Comparison of the MSE in estimating the drift coefficient, $\mathcal{L}_\mu$, defined in \eqref{eq:orthog_2}, between the proposed method with the benchmark of \cite{duncker2019learning}, across different datasets.}
		\label{tab:sde_results_duncker}
	\end{table*}
}

\section{Comparison with \cite{duncker2019learning}}
\label{sec:comparison_duncker2019learning}
We add a comparison where we apply the method proposed in \cite{duncker2019learning} to the true latent SDE.
Since \cite{duncker2019learning} has the advantage of observing the true latent SDE realization, rather than the high dimensional ambient space observations, the method should act as a lower bound to the proposed method. 
The results are reported in Table~\ref{tab:sde_results_duncker} along with the Cram\'er-Rao lower bounds previously obtained and $p$-values on a two sample $t$-test between the statistics of the two methods. 
In this case, we see that the method in \cite{duncker2019learning} performs very well in low dimensional cases with easier SDEs (namely the 1D and 2D OU and the constant) but does worse in the other cases. 
Moreover, when considering the statistical significance of the differences between the methods, the method by \cite{duncker2019learning} exhibits statistical significance in the low dimensional OU processes (for $p < 0.01$) and in the 4D constant case (for $p < 0.05$). 
The proposed method exhibits statistical significance for the Circle and Double Well experiments (for $p < 0.01$).
This provides greater evidence of the efficacy of the proposed method since, with the much more difficult estimation task from the ambient space, the proposed method performs within one order of magnitude in error in all cases and better with higher dimensional latent space.

\section{Comparison with image registration techniques}
\label{sec:comparison_image_registration}

\begin{table}[]
	\centering
	\small
	\begin{tabular}{llll}
		DNA Dataset & $V= 0$ & $V=1$ & $V=2$ \\
		\toprule 
		ITK & $2.58 \times 10^0$ & $4.27  \times 10^0$ & $8.28 \times 10^0$ \\
		Proposed & $7.93 \times 10^{-2}$ & $1.87 \times 10^{-1}$ & $2.03 \times 10^{-1}$
	\end{tabular}
	\caption{MSE between estimated displacements and approximate coordinates estimated by intensity segmentation. }
	\label{tab:registration}
\end{table}

We add a comparison between our method and image registration techniques in estimating the molecules displacement in the Fluorescent DNA datasets. We used the \texttt{TranslationTransform} method in the Insight ToolKit (ITK) \cite{avants2014insight} to estimate the displacement between frames. 
The results of the MSE between the estimated displacement from the registration algorithm and the coordinates approximated from the intensity based particle tracking are presented in Table~\ref{tab:registration}.
We observe that the proposed method outperforms the image registration method in the DNA datasets. We believe this may be due to noise, which makes tracking the position of the molecule difficult. Additional preprocessing may be necessary to achieve better results with image registration techniques.

\end{document}